\pdfoutput=1
\documentclass[11pt]{article}

\usepackage[hyphens]{url}
\usepackage{hyperref}
\usepackage{xr}
\usepackage[preprint]{acl}

\usepackage{times}
\usepackage{latexsym}

\usepackage[T1]{fontenc}
\usepackage[utf8]{inputenc}

\usepackage{microtype}

\usepackage{inconsolata}

\usepackage{graphicx}

\usepackage{amsmath}
\usepackage{amssymb}
\usepackage{booktabs}
\usepackage{tcolorbox}
\usepackage{multicol}

\usepackage{pifont}
\usepackage{bbding}
\usepackage{fontawesome}
\usepackage{subcaption}
\usepackage{multirow}
\usepackage{float}

\usepackage{listings} % 加载 listings 宏包

\title{Synthesis by Design: Controlled Data Generation via Structural Guidance}
\author{
 \textbf{Lei Xu\textsuperscript{1}},
 \textbf{Sirui Chen\textsuperscript{1, 2}},
 \textbf{Yuxuan Huang\textsuperscript{1,3}},
 \textbf{Chaochao Lu\textsuperscript{1}\thanks{Corresponding author.}}
\\
\\
 \textsuperscript{1}Shanghai Artificial Intelligence Laboratory,
 \textsuperscript{2}Tongji University,
 \textsuperscript{3}Shanghai Jiao Tong University
\\
 \texttt{\{xulei, chensirui, huangyuxuan, luchaochao\}@pjlab.org.cn}
}

\begin{document}

\newcommand{\cl}[1]{{\color{red}\textbf{\fontsize{8}{8}\selectfont [CL: #1]}}}

\newcommand{\lx}[1]{{\color{blue}\textbf{\fontsize{8}{10}\selectfont [LX: #1]}}}

\newcommand{\src}[1]{{\color{teal}\textbf{\fontsize{8}{10}\selectfont [SRC: #1]}}}

\maketitle
\begin{abstract}
Mathematical reasoning remains challenging for LLMs due to complex logic and the need for precise computation. 
Existing methods enhance LLM reasoning by synthesizing datasets through problem rephrasing, but face issues with generation quality and problem complexity.
To address this, we propose to  extract structural information with  
generated problem-solving code from mathematical reasoning and guide data generation with structured solutions. 
Applied to MATH and GSM8K, our approach produces 39K problems with labeled intermediate steps and a 6.1K-problem benchmark of higher difficulty.
Results on our benchmark show that model performance declines as reasoning length increases. Additionally, we conducted fine-tuning experiments using the proposed training data on a range of LLMs, and the results validate the effectiveness of our dataset. We hope the proposed method and dataset will contribute to future research in enhancing LLM reasoning capabilities. 
Our code and data are available at: \url{https://github.com/OpenCausaLab/StructuralGeneration}.
\end{abstract}

\section{Introduction}
In recent years, large language models (LLMs) have been developed at a rapid pace and have shown great improvement in natural language understanding, generation, and problem-solving abilities. 
Their success has made them a popular choice for various applications, including mathematical problem-solving  \cite{CSV, MATHCODER, PAL}, code generation  \cite{nijkamp2023codegen, chen2021codex}, and open domain conversation  \cite{openai2024gpt4technicalreport, ouyang2022traininglanguagemodelsfollow}.
Despite their impressive performance, LLMs still struggle with reasoning tasks  \cite{valmeekam2023planningabilitieslargelanguage, CSV}. The most representative task is solving mathematical problems, which requires complex and multi-step reasoning capabilities, while LLMs are short of these abilities and often produce nonsensical or inaccurate content  \cite{STEPWISEDPO}.

% Step数据的重要性，R1的成功也同样需要Step数据，但是它难以获得. PRM800K，
% 这段还需要改一下，需要把对step信息的利用放进来
% To address this challenge, previous works resort to prompting-based approaches such Chain-of-Thougth [ref], which enhances LLM's reasoning ability by asking them to generate intermediate steps in their reasoning process. 
To address this challenge, post-training has proven effective in enhancing LLMs' reasoning abilities. Previous research  \cite{MAMMOTH, zeng2023agenttuningenablinggeneralizedagent} shows that fine-tuning pre-trained LLMs on high-quality Chain-of-Thought (CoT) reasoning data improves performance by enabling models to better leverage pre-trained knowledge. A notable example is GPT-o1  \cite{GPTO1}, which introduced inference-time scaling by extending the length of the CoT reasoning process. 

% Step数据的重要性，R1的成功也同样需要Step数据，但是它难以获得. PRM800K，
% 人工成本高昂，导致Step数据十分稀少。因此需要合成数据
Although post-training offers valuable dense supervision, the scarcity of high-quality CoT reasoning data remains a major challenge. To address this, human-annotated datasets like PRM800K \cite{PRM800K} have been introduced. However, such data are expensive and hard to scale, making automatic data generation a more practical alternative.
Various data augmentation techniques  \cite{MAMMOTH,tang2024mathscalescalinginstructiontuning, METAMATH, WIZARDMATH, shao2024deepseekmathpushinglimitsmathematical} have been proposed to improve LLMs' reasoning abilities.
For example, MetaMath  \cite{METAMATH} employs a bootstrapping method to rephrase mathematical questions from multiple perspectives.
% Wizzard Math  \cite{WIZARDMATH} introduces Evlo-instruct to create questions with varying complexity. 
% they cannot guarantee the correctness of intermediate reasoning steps, nor can they obtain data with process supervision information, which may lead to hallucinations.
 \cite{MATHCODER} integrated the code into the reasoning process, which allows models to self-verify and correct their reasoning through execution results.

As noted by \cite{saxton2019analysingmathematicalreasoningabilities}, solving mathematical problems relies on two core abilities: linguistic comprehension and mathematical reasoning. Existing data generation methods primarily focus on the former, with limited attention to the structural aspects of reasoning. As a result, they often face the following issues:
\begin{enumerate}
    \item \textbf{Label Missing}: Without access to the intermediate computational process, generating accurate answers for new problems is challenging \cite{MATHCODER} for existing methods, let alone providing step-level supervision.
    \item \textbf{Agent-Dependent Quality}: The quality of both reasoning paths and answers depends heavily on the agent, making the generation process vulnerable to issues like hallucinations  \cite{STEPWISEDPO} and reward hacking  \cite{skalse2022definingcharacterizingrewardhacking, REFT}.
    \item \textbf{Limited Data Expansion}: Constrained by the agent’s reasoning ability, existing methods struggle to generate problems beyond what the agent can solve.
\end{enumerate}

\begin{figure*}[h]
    \centering
    \includegraphics[width=0.9\linewidth]{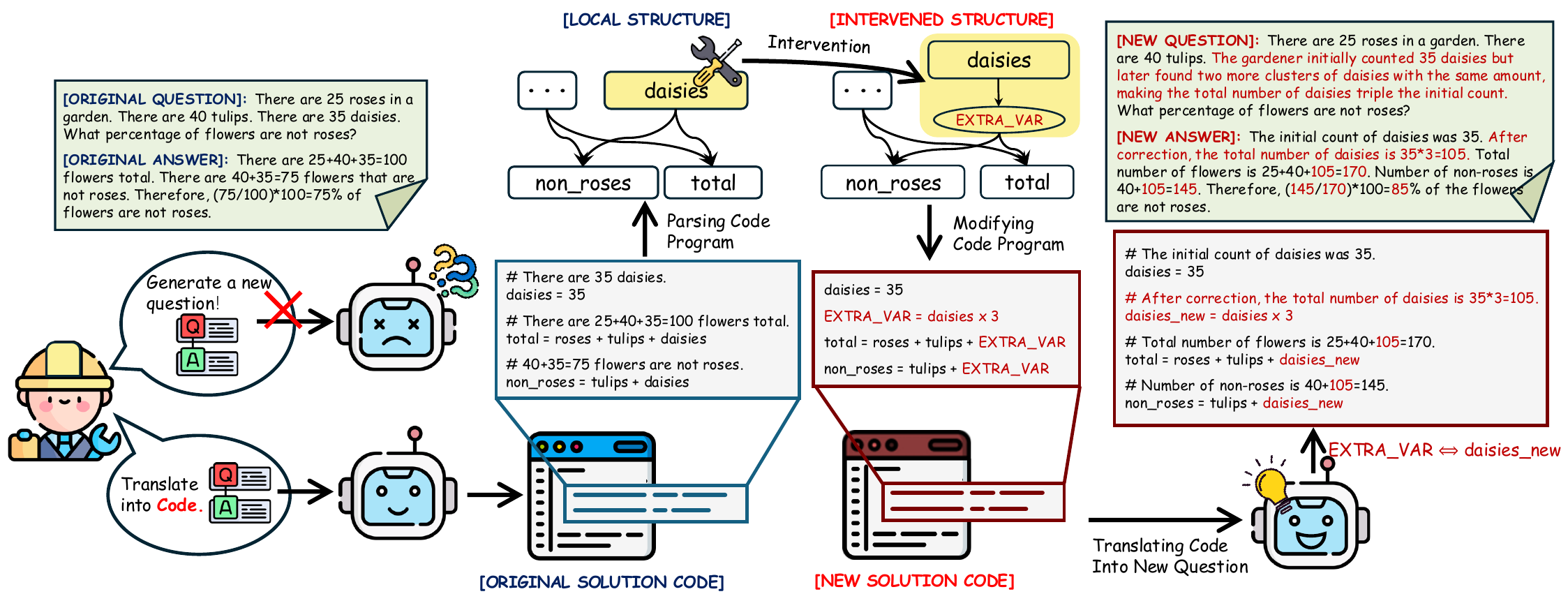}
    \caption{Illustration of the proposed data generation framework. Given the original problem and its reasoning process, we first ask the agent to translate the answer into solution code, where each line of code corresponds to one reasoning step. Then, we parse the code program to extract the computational structure of the original problem, and intervene on this structure to generate solution code for a new problem. We provide the new code to the model and ask it to generate a new problem based on the context of the original problem.}
    \label{fig:main}
\end{figure*}

In this work, we emphasize the role of structural information in mathematical reasoning and use it to generate structured reasoning data.
As shown in Figure~\ref{fig:main}, starting from existing problems, we translate their reasoning processes into executable Python code, which simplifies semantics and constrains reasoning within a computational structure.
The program is composed of code-reasoning pairs, where each line computes an intermediate result and is commented with its corresponding reasoning step. 
This provides both ground truth and step-level supervision (addressing Problem 1), while also capturing dependencies between steps. 
Leveraging this structure, we generate new problems by modifying the code with minimal changes.
% Based on this property, we can generate new reasoning processes by modifying the code structure, allowing the agent to generate new problems with minimal changes to the original code.
% With this property at hand, we can easily generate new reasoning processes by modifying the code structure (as illustrated in Figure X). The modified code is then provided to the agent along with the original problem statement, due to the minimal modifications between the original and new code, the agent can naturally translate these changes into the original problem’s context and generate new problems. 

% Through this approach, the intervention in mathematical problems is decomposed into two tasks: first, local intervention on the program, which ensures strict control over the correctness of intermediate computational steps (addressing Problem 2); and second, translation by the agent, which converts the modified code into a semantically meaningful problem and reasoning process. Since the model has access to contextual information and the correct answers beforehand, the problem generation process is transformed into a translation task rather than a reasoning task, preventing the model from relying on its own reasoning ability and ensuring that the difficulty of generated problems is no longer constrained by the agent’s reasoning capability (addressing Problem 3).

The proposed method decomposes the task of math problem generation into two tasks: first, local intervention on the code to ensure correctness of intermediate steps (addressing Problem 2), and second, translation by the agent to convert the modified code into a coherent problem and reasoning process. With access to context and answers, the problem generation becomes a translation task rather than a reasoning one, reducing reliance on the model’s reasoning ability and allowing problem difficulty to exceed the agent’s capabilities (addressing Problem 3). 

\textcolor{black}{Moreover, by leveraging structural information, we decompose solutions into steps and quantify problem complexity by step count. While prior work  \cite{qi2025quantifying} shows that complexity affects model performance, few studies offer quantitative ways to measure it or provide datasets with such annotations. Our method addresses this gap by analyzing and annotating real-world datasets with step information.}

The contributions of this work are summarized as follows:
\begin{itemize}
    \item From a \textbf{methodological} perspective, we propose a new approach that uses structural information to monitor, guide, and intervene in mathematical reasoning. This structure enables the generation of intervention data with both step-level and outcome-level supervision. \textcolor{black}{Importantly, it allows us to  quantify problem complexity by step count.}
    \item From a \textbf{data} perspective, we create more challenging datasets based on MATH and GSM8K. Unlike existing methods, our datasets are derived from executable code, ensuring a more reliable computational process. 
    \item From a \textbf{model} perspective, we finetune Qwen-2-7b, Mistral-7b, and LLaMA-3-8b series models on our datasets. Experimental results on MATH and GSM8K show the superior performance of our data.
\end{itemize}

%[TODO待看]
\section{Related Work}
\subsection{LLM for Mathematical Reasoning}
Various methods have been proposed to improve the mathematical reasoning abilities of LLMs, generally, they fall into two categories: (1) prompting-based and (2) finetuning-based methods. Prompting-based methods use carefully designed prompts to guide reasoning without altering model parameters. For instance, Chain-of-Thought (CoT)  \cite{COT} prompts models to break problems into sub-steps, generating intermediate reasoning paths toward the answer. 
Finetuning-based methods improve open-source LLMs via instruction tuning, often leveraging data augmentation to create effective training data. Rejection Sampling Finetuning (RFT)  \cite{RFT} selects optimal reasoning paths from candidates using a trained reward model. MetaMath  \cite{METAMATH} rewrites questions from multiple perspectives to expand training data. WizardMath  \cite{WIZARDMATH} applies reinforced evol-instruct techniques to generate new questions from GSM8K and MATH, then uses them for SFT and PPO training to improve reasoning. Existing methods typically follow a ``Problem $\rightarrow$ Answer'' workflow, which relies on the LLM to produce the final answer. However, for complex questions, limited reasoning ability may lead to unreliable outputs. In contrast, our finetuning-based approach adopts a reversed ``Answer $\rightarrow$ Problem'' workflow to generate reliable step-by-step reasoning data.

\subsection{Code-integrated Data Generation}
Since LLMs are known to be less reliable in arithmetic computations  \cite{PAL, lewkowycz2022solving}, some studies have attempted to incorporate code for improved accuracy. For example, GSM8K dataset  \cite{GSM8K} embeds code within solutions to obtain intermediate results. The CSV method  \cite{CSV} uses prompt-based code generation to compute and verify intermediate steps. Building on this idea, Wang et al. introduced the MathCodeInstruct dataset  \cite{MATHCODER}, which includes natural language, code, and execution outputs for each problem, enabling finetuning to improve reasoning. However, few approaches have explored or leveraged the structural information inherent in mathematical problems.
 \textcolor{black}{In this regard, GSM-Symbolic  \cite{gsmsymbolic} extracts and modifies computational information in problems via pattern matching. Program-Aided Language model (PAL)  \cite{PAL} generates code through prompts as intermediate steps and computes final answers by executing the code. While such methods can create more challenging problems by altering variable values, the lack of alignment between code and reasoning steps may lead to implausible counterfactuals and hinder access to intermediate results.}
In addition to this,  \citet{xie2024adversarialmathwordproblem} obtained this structural information by parsing code programs and used abstract syntax trees to structurally generate adversarial examples for model evaluation. 
Building on code-based approaches, our work fully leverages the implicit dependencies in programs to generate accurate intermediate steps and construct more challenging problems by extending the code structure.

\section{Method}
We construct our dataset based on MATH  \cite{MATH} and GSM8K  \cite{GSM8K} datasets. As shown in Figure~\ref{fig:main}, our method consists of two stages. First, we instruct the agent to translate the original reasoning process into code program, where the corresponding reasoning process is annotated before each code block. Based on this, we intervene in the code and translate the modified program into new problems and reasoning processes. Furthermore, we introduce a multi-level verification mechanism to control the quality of the generated data. 
% To verify the quality of the proposed dataset, we employ the step-controlled DPO [ref] strategy to finetune LLM with our datasets. Next, we will introduce these steps sequentially.

\subsection{Structure Extraction from Reasoning Process}
\label{sec:structure_extract}

For a given problem $ q $ and its corresponding solution process $ s = (t_1, t_2, ..., t_m) $, where $ t_i $ represents a natural language step, we ask the agent to translate the reasoning process into a solution program, which interleaves step-by-step reasoning with code blocks. The prompt for code generation can be found in Appendix \ref{appen:code_gen}.

Based on this, we parse the program and represent it as a tree-like computational graph\footnote{We use a tree analogy for ease of explanation, despite the structure allowing some deviations from a strict tree definition.}, as shown in Figure~\ref{fig:main}. In this graph, the leaf nodes represent the variable mentioned in the problem, the root node represents the final answer to the problem, and the intermediate nodes are related to different reasoning steps.

It is worth noting that while our computational structure resembles an abstract syntax tree (AST) \cite{xie2024adversarialmathwordproblem}, it differs by replacing operation nodes with intermediate variables. This design allows us to capture the dependencies between variables and understand how each contributes to the final result. For instance, in Figure~\ref{fig:main}, the variable \verb|daises| influences both \verb|non_roses| and \verb|total|. Modifying \verb|daises| thus affects these variables and ultimately alters the final output.

\subsection{Structural Intervention}
\label{sec:structural_interv}
Based on the structure obtained in Section~\ref{sec:structure_extract}, the simplest way to generate new problems is to modify the leaf nodes (variables in the problem), and then obtain new intermediate computational steps and the final result through the computational structure. It is worth noting that although PAL  \cite{PAL} constructs new problems in a similar manner, our approach explicitly associates the intermediate reasoning steps with code, allowing us to easily generate new computational steps through pattern matching.

Although the above intervention method produces new problems, it still essentially relies on the original computational structure. Prior works  \cite{saxton2019analysingmathematicalreasoningabilities, qi2024quantifyinggeneralizationcomplexitylarge} have pointed out that long reasoning problems significantly impact the performance of LLMs. From the training perspective, long reasoning problems can also improve the generalization ability of the model  \cite{saxton2019analysingmathematicalreasoningabilities}. Considering these factors, we propose leveraging the computational structure to construct harder math problems that need longer reasoning steps.

Specifically, given a graph structure $ G(T,E) $ with $ T = \{t_1,t_2,\dots,t_m\} $ representing variable nodes corresponding to each reasoning steps and $ E $ representing variable relationships, we first randomly select a node $t_i$ from the leaf nodes and intermediate nodes, and generate a proxy node $t_i'$ that is mapped from $ t_i $. Namely, $ t_i' = f(t_i) $, where the function $f$ can be seen as the intervention on $t_i$. As shown in Figure~\ref{fig:main}, we introduce a proxy variable \verb|EXTRA_VAR| and replace \verb|daises| by this new variable for the subsequent computations.
% \footnote{It is noteworthy that the structural intervention is not a straightforward approach as it seems, in Appendix \ref{appen:proxy}, we describe the potential problem it may cause and how we handle this problem so that we can intervene on any node in the structure.} 
Due to the sequential nature of programming languages, all computations involving \verb|daises| will appear after the modified line of code. Thus, this intervention can be implemented easily by replacing all occurrences of $ t_i $ with $ t_i' $ in the subsequent code. 

To obtain the mapping function $ f $, since GSM8K problems only involve grade school arithmetic, we set $ f $ to simple addition, subtraction, or multiplication with a random value. Note that division is excluded from the interventions as it often results in non-intuitive or counterfactual outcomes (e.g., ``0.3 persons''). Since MATH problems involve much more complex and domain-specific mathematical operations, we allow an agent to decide how to introduce new reasoning steps. Further details can be found in Appendix \ref{appen:code_gen}.

Based on the modified structure, we ask the agent to translate it into a new reasoning process and generate a corresponding question. During this process, the original question and the solution code are also provided to the agent as contextual information. Since the modified program only slightly differs from the original, the agent needs to adapt only the changed parts. To guide accurate computation, we annotate each line of code with its execution result as ground truth. The prompts for reasoning and problem generation are provided in Appendix~\ref{appen:code_trans}.

\subsection{Quality Control}
We adopt a multi-level validation approach to ensure the quality of intervention data, including (1) local code execution and (2) agent evaluation.

For local validation, we compare the execution results of the generated code with the ground truth from the original text to verify the correctness of the code produced in Section~\ref{sec:structural_interv}. For structural interventions, we compare the execution results of each variable in the modified code with the original results and filter out cases where numerical sign inversion occurs (e.g., a positive number becomes negative) or data type changes (e.g., an integer becomes a floating point).

Despite providing the groundtruth for the reasoning process, large language models (LLMs) may still generate counterfactual outputs or produce mismatched questions and reasoning processes. To mitigate this issue, in agent evaluation, we first ask the agent to perform self-evaluation during the generation process. Additionally, to prevent overconfidence in its own outputs (i.e., the situational awareness issue  \cite{berglund2023takencontextmeasuringsituational}) and reduce cognitive biases introduced by a single model, we employ an external model from a different family to assess the generated results. Only data that passes both self-evaluation and external agent evaluation will be retained. In this study, we use the Claude-3-5-Sonnet model for data generation and GPT-4o for evaluation. Although our approach does not impose strict requirements on the models used, we found that Claude-3-5-Sonnet produces semantically richer questions compared to GPT-4o. 
% A comparative analysis of questions generated by these two models is provided in Appendix \ref{appen:agent_diff}.

% 这块具体的做法待补充
We also conduct a human evaluation to evaluate and assess the quality of the generated data, the detailed information is provided in Appendix \ref{appen:quality_analysis}.

\begin{table}[!t]
    \centering
    \scalebox{0.8}{
    \begin{tabular}{ccccc}
\toprule
Dataset   & GSM8K$_{tr}$  & MATH$_{tr}$  & GSM8K$_{te}$  & MATH$_{te}$ \\ \midrule
Code Gen. & 7.1K        & 6.0K         & 1.3K        & 3.9K        \\\midrule
Extension & 14.3K       & 11.6K        & 3.2K        & 2.9K        \\ \midrule
Total     & \multicolumn{2}{c}{39.0K} & \multicolumn{2}{c}{11.3K} \\ \bottomrule
\end{tabular}
}
    \caption{Number of generations in the proposed dataset, where $tr$ denotes training set and $te$ denotes testing set, ``Code Gen'' represents samples directly translated from the original reasoning process and ``Extension'' represents samples obtained after structural expansion. }
    \label{tab:stat_dist}
\end{table}

\section{Experiments}
In this section, we first present the datailed information of the proposed dataset. Then we test state-of-the-art
LLMs on the proposed benchmark. Finally, we demonstrate the effectiveness of our data by showing the fine-tuning results on Mistral, Qwen, and Llama series models.
\subsection{Dataset Analysis}
\paragraph{Data Generation}
We generate our data using two widely used math datasets: (1) GSM8K  \cite{GSM8K}, a grade school mathematics dataset containing 7,473 training and 1,319 test samples, and (2) MATH  \cite{MATH}, which contains 7,500 training and 5,000 test samples covering high school competition-level problems across five difficulty levels and seven subjects. Both datasets provide full solution processes, which facilitate the code generation process of our method.
\paragraph{Dataset Statistics}
Table~\ref{tab:stat_dist} shows the number of samples generated from GSM8K and MATH. We apply progressive expansion by intervening based only on previously successful generations. For the training sets GSM8K$_{tr}$ and MATH$_{tr}$, each sample undergoes up to three rounds of intervention, while for the test set, we allow up to eight rounds. If no valid sample is produced after intervention on a given question, we stop further attempts on that question. This process is repeated for three times, resulting in a total of 50K question-answer pairs with intermediate reasoning steps. Among these, 39K samples (including ``Code Gen'' \& ``Extension'') from the original training set are used for model training, while 6.1K ``Extension'' samples from the test set are used as a testing benchmark to evaluate model performance on more challenging data. We term our benchmark as ``MATH-Struct'' and ``GSM8K-Struct''.

\begin{figure}[!t]
    \centering
    \begin{subfigure}[t]{0.23\textwidth}
        \centering
        \includegraphics[width=1\linewidth]{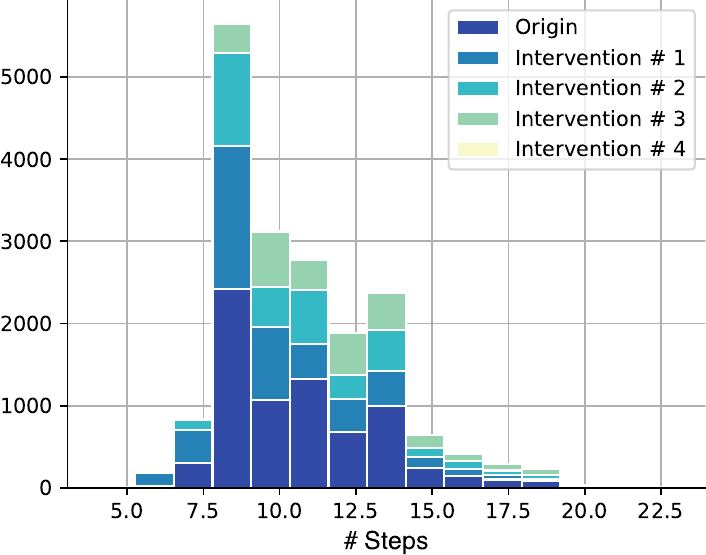}
        \caption{Distribution of GSM8K$_{tr}$.}
        \label{fig:steps_sub1}
    \end{subfigure}
    \hfill  
    \begin{subfigure}[t]{0.23\textwidth}
        \centering
        \includegraphics[width=1\linewidth]{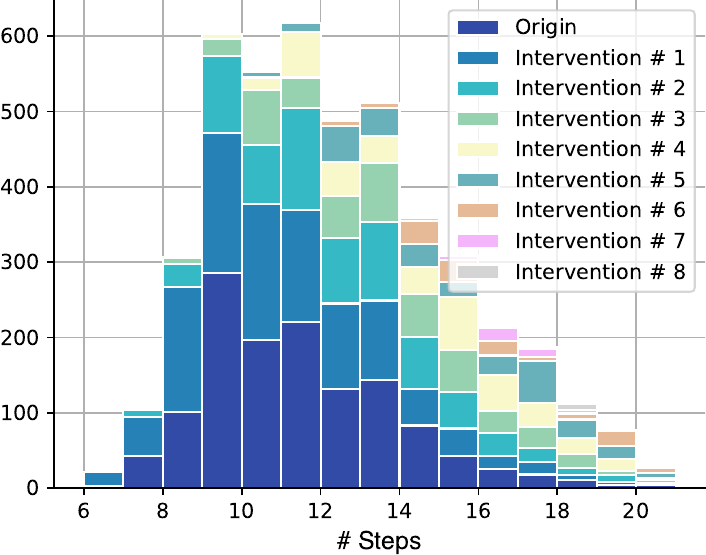}
        \caption{Distribution of GSM8K$_{te}$.}
        \label{fig:steps_sub2}
    \end{subfigure}
    \hfill  
    \begin{subfigure}[t]{0.23\textwidth}
        \centering
        \includegraphics[width=1\linewidth]{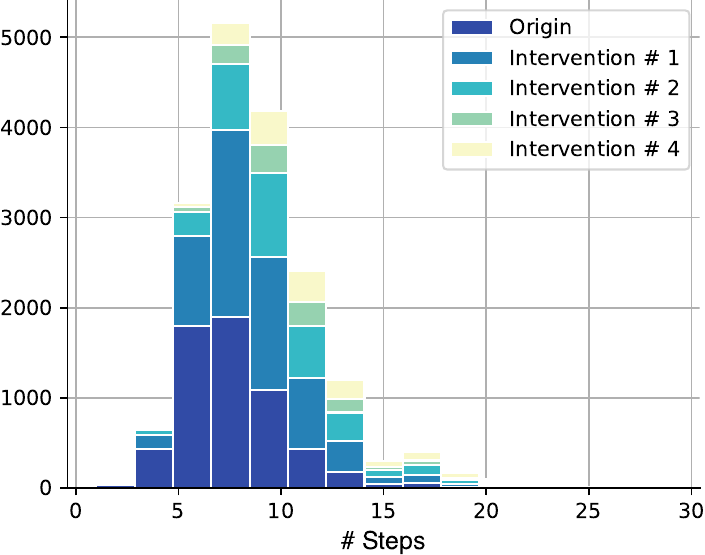}
        \caption{Distribution of MATH$_{tr}$.}
        \label{fig:steps_sub3}
    \end{subfigure}
    \hfill  
    \begin{subfigure}[t]{0.23\textwidth}
        \centering
        \includegraphics[width=1\linewidth]{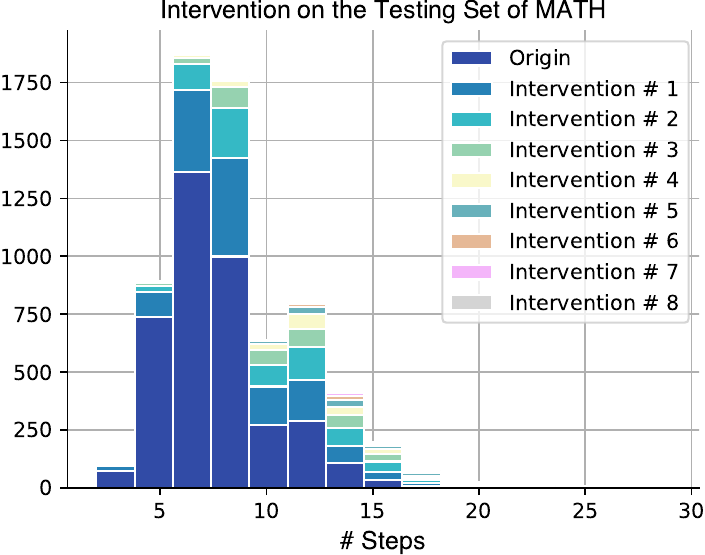}
        \caption{Distribution of MATH$_{te}$.}
        \label{fig:steps_sub4}
    \end{subfigure}
    \caption{Step distribution across different data components.}
    \label{fig:step_dist}
\end{figure}

\begin{figure*}[!htbp]
    \centering
    \includegraphics[width=0.9\linewidth]{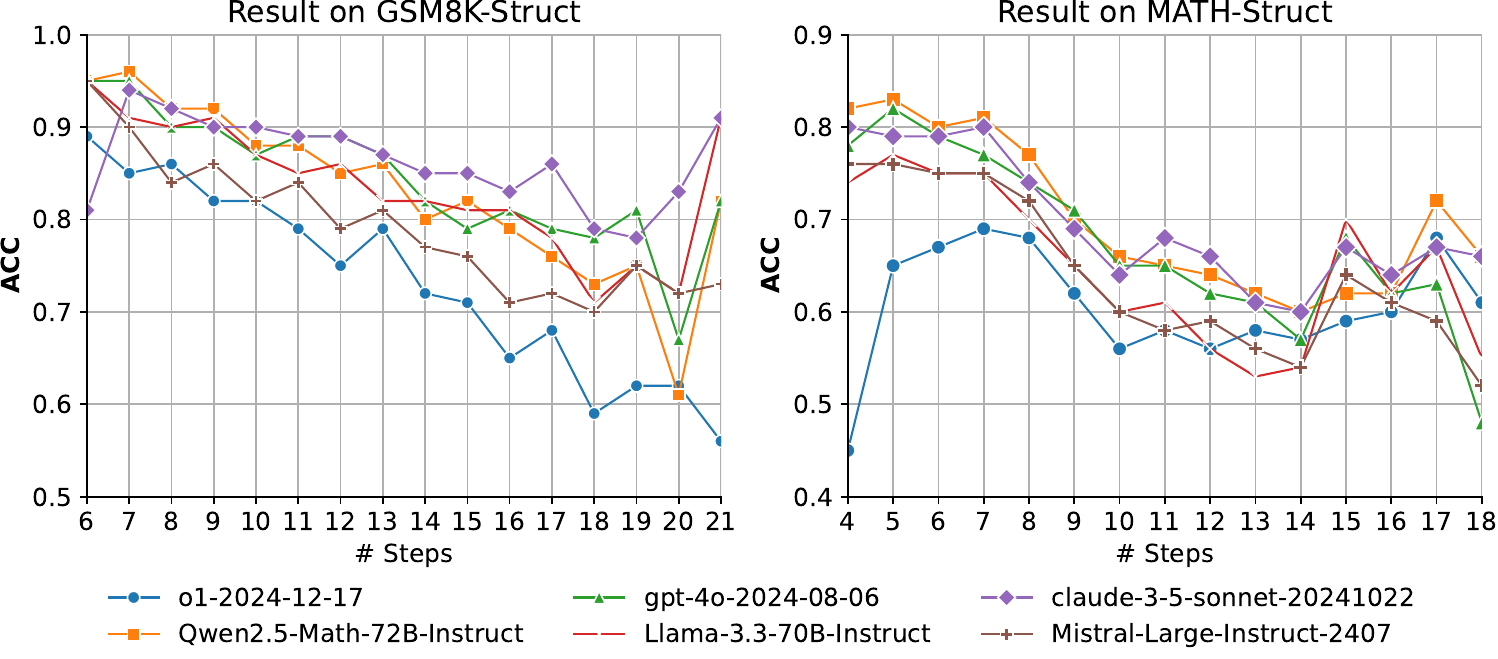}
    \caption{Benchmarking results on different steps.}
    \label{fig:results_steps}
\end{figure*}

% \begin{table}
%     \centering
%     \scalebox{0.75}{
%     \begin{tabular}{|c|c|c|c|c|c|c|c|c|}
%     \hline
%        \# Inteventions  & 1 & 2 & 3 & 4 & 5 & 6 & 7 & 8\\
%    \hline
%    \hline
%         GSM8K & 1101 & 734 & 479 & 390 & 285 & 130 & 46 & 11 \\\hline
%         MATH & 1379 & 750 & 386 & 220 & 117 & 62 & 32 & 14 \\
%     \hline
%     \end{tabular}
%     }
%     \caption{Distribution of generated data with increasing round of interventions.}
%     \label{tab:stat_steps}
% \end{table}

\paragraph{Difficulty Distribution}
Our method decomposes the overall reasoning process into individual steps. As noted in \cite{qi2024quantifyinggeneralizationcomplexitylarge}, problem difficulty is influenced by the intrinsic complexity of the task, which is reflected in the number of reasoning steps required. Based on this, we analyze the difficulty distribution of both original and intervened samples by measuring the number of steps in each problem.

Specifically, after obtaining the solution code for a given problem, we count the number of code blocks at the first level of indentation as an approximation of the solving steps. Using this measure, we plot frequency distribution histograms for the number of solving steps in both the original data and the intervened data after each intervention, the results are shown in Figure~\ref{fig:step_dist}.

Overall, the step distribution after intervention shifts to the right compared to the original distribution, indicating an increase in dataset difficulty. Further comparison between the original and intervened distributions reveals that most intervened data points are located on the right side of the original distribution. This suggests that our intervention method generates more problems with longer solving steps.

Examining the trend from intervention round \#1 to intervention round \#8, we observe that as the number of generation round increases, the problems become more complex and difficult. This increased complexity makes it more challenging for the model to comprehend the problems, leading to a decrease in the amount of successfully generated data.
\begin{table*}[!htbp]
 \centering
    \resizebox{0.88\linewidth}{!}
{
\begin{tabular}{ccccccccc}
\toprule
                            & Dataset & Baseline         & MetaMath                         & MathCoder & Math-shepherd & MammoTH                           & MathScale                        & Ours                              \\ \midrule
\multirow{2}{*}{Mistral-7b} & GSM8K   & 8.72\%           & \underline{60.65\%} & 40.26\%   & 40.41\%       & 60.20\%                           & 48.75\%                          & \textbf{68.31\%}                  \\
                            & MATH    & 4.50\%           & 16.46\%                          & 12.30\%   & 16.58\%       & 15.34\%                           & \textbf{18.62\%}                 & \underline{16.90\%} \\ \midrule
\multirow{2}{*}{Qwen-2-7b}  & GSM8K   & 67.55\%          & 69.83\%                          & 61.87\%   & 44.73\%       & \underline{74.68\%} & 64.37\%                          & \textbf{78.01\%}                  \\
                            & MATH    & \textbf{46.20\%} & 27.66\%                          & 31.16\%   & 23.94\%       & \underline{40.78\%} & 39.16\%                          & 39.70\%                           \\ \midrule
\multirow{2}{*}{Llama-3-8b} & GSM8K   & 7.13\%           & \textbf{72.18\%}                 & 46.25\%   & 41.85\%       & 59.51\%                           & 45.64\%                          & \underline{71.34\%}  \\
                            & MATH    & 4.66\%           & \textbf{23.68\%}                 & 16.94\%   & 19.30\%       & 18.26\%                           & \underline{21.58\%} & 19.70\%                           \\ \midrule
\multicolumn{2}{c}{\# 1st}            & 1                & 2                                & 0         & 0             & 0                                 & 1                                & 2                                 \\ \midrule
\multicolumn{2}{c}{\# 2nd}            & 0                & 1                                & 0         & 0             & 2                                 & 1                                & 2                                 \\ \bottomrule
\end{tabular}
}
	\caption{Performance on original GSM8K and MATH datasets, where ``Baseline''s refer to the pretrained version of LLMs. The best results are in \textbf{bold} and second best are \underline{underlined}.}
\label{table:result_on_origin}
\end{table*}
\subsection{Benchmarking Result}
To clearly understand the performance of various models on problems of different difficulty levels, we conducted extensive evaluations on both the original problems and their extended versions using state-of-the-art models. 
\textcolor{black}{For better illustration, we select and evaluate representative models in the main text, }including \verb|GPT-o1|\footnote{For the sake of cost efficiency, we only evaluated the performance of GPT-o1 on the extended dataset.}  \cite{GPTO1}, \verb|GPT-4o|  \cite{GPT4O}, \verb|Claude-3-5-Sonnet|  \cite{CLAUDE35}, \verb|Qwen2.5-MATH-Instruct|  \cite{QWEN25}, \verb|Llama-3.3|  \cite{grattafiori2024llama3herdmodels}, and \verb|Mistral-Large-Instruct|  \cite{MISTRALLARGE}.
We evaluate more models and provide their results in Appendix \ref{appen:add_benchmark_results}.

We adopted the prompt format used by MathCoder\footnote{\url{https://github.com/mathllm/MathCoder/blob/main/src/inference.py}}, requiring models to enclose their answers within the \verb|\boxed{}| tag. All models were tested in a zero-shot setting, and the accuracy based on one single generation is used as the evaluation metric.

Figure~\ref{fig:results_steps} presents the performance of different models across problems of varying difficulty levels\footnote{Due to the scarcity of high-difficulty questions (i.e., GSM8K with more than 21 steps and MATH with more than 18 steps), which makes it difficult to reliably assess model performance, we omit these results to better highlight the overall performance trend.}.
Overall, as the number of reasoning steps increases, the performance of all models declines, which supports the existing opinion that longer reasoning steps correspond to more challenging problems. As shown in Figure~\ref{fig:step_dist}, there are fewer samples with more than 15 steps in the MATH dataset, leading to greater performance fluctuations across models in this range.

Since Claude-3-5-Sonnet and GPT-4o were involved in the data generation process, the generated data inevitably carries some of their biases. Claude-3-5-Sonnet and GPT-4o tend to retain problems they can understand during data generation, which contributes to their overall superior performance. However, it is important to emphasize that during the generation phase, the models rely solely on their understanding and creativity, as they are provided with supervision information for each step. In contrast, during the evaluation phase, without such guided information, their performance depends on their reasoning abilities.

It is worth noting that the performance of the GPT-o1 model was lower than our expectations. In Appendix~\ref{appen:analysis_of_o1}, we examined the samples where GPT-o1 produced incorrect answers, we found that GPT-o1 tends to overlook key information and makes simple calculation errors during long reasoning processes. Since o1 is reported to suffer from a ``dumbing down'' issue\footnote{\url{https://community.openai.com/t/is-anyone-elses-gpt-4o-and-o1-suddenly-acting-dumb}}, it is important to note that the current evaluation results may not accurately reflect o1's true performance.

In Appendix~\ref{sec:res_on_interv_num}, we evaluate model performance on datasets generated with different numbers of intervention rounds. We observe that performance declines as the number of rounds increases, indicating that additional interventions result in more challenging problems. To address potential quality issues from automated generation, in Appendix~\ref{sec:res_on_filtered}, we focus on the subset of samples correctly reasoned by both Claude-3.5-Sonnet and GPT-4o. We then evaluate all models on the union of these validated samples. As shown in Appendix~\ref{sec:res_on_filtered}, the performance trends on this subset are consistent with those in Figure~\ref{fig:results_num_interventions}, which further confirms the reliability of our data.

\subsection{Finetuning Results}
In this section, we use the generated data for model training to verify the efficacy of the generated data in improving the model's reasoning performance.
\paragraph{Comparing Methods}
According to Table~\ref{tab:stat_dist}, we selected a total of 39K data from GSM8K$_{tr}$ and MATH$_{tr}$, including both Code Generation and Extension tasks, for model training. We compared the MathCoder  \cite{MATHCODER}, MetaMath  \cite{METAMATH}, Math-Shepherd \cite{mathshepherd}, MammoTH \cite{MAMMOTH}, and MathScale \cite{mathscale} datasets. For a fair comparison, we selected 39K data from each dataset to match the size of our dataset. Since MetaMath consists of four categories: `\verb|AnsAug|', `\verb|FOBAR|', `\verb|Rephrase|', `\verb|Self-Verification|', for each category, we selected 4875 samples evenly. For other datasets, we randomly selected 39K samples. 

\paragraph{Training Details}
We selected four baseline SFT models: \verb|Mistral-7b-v0.1|  \cite{MISTRAL7B}, \verb|Qwen-2-7b|  \cite{QWEN27B}, and \verb|Meta-Llama-3-8b|  \cite{grattafiori2024llama3herdmodels}
% , and \verb|Llama-3-70b|  \cite{grattafiori2024llama3herdmodels}
. Using the Llamafactory training framework  \cite{LLAMAFACTORY}, we trained the models on 8 A100 GPUs. 
% ollowing the approach in  \cite{METAMATH}, for the Llama-3-70b model, we used QLoRA for fine-tuning, setting the LoRA rank and alpha to 96 and 16, respectively, with a dropout rate of 0.05. 
% For the other models, we conducted full hyperparameter tuning.
We set the sequence length to 2048 and perform full parameter tuning on each model for two epochs. 
% For Llama-3-70b, the batch size was set to 16, while 
For all models, the batch size was set to 256. The learning rate for all models was chosen from the set \verb|{1e-5, 2e-5, 2e-6}|. 
Since both MathCoder and our dataset dataset contain Python code, we introduced two special tokens, <|\verb|start_code|> and <|\verb|end_code|>, to clearly distinguish code from text content during fine-tuning with these datasets. 
For the other models, according to the original paper  \cite{METAMATH}, we employ the Alpaca-style  \cite{alpaca} prompt for training.
We provide the training prompt in Appendix \ref{appen:prompt_for_training}.

\subsection{Results on Original Dataset}
We first evaluated models trained on different datasets using the original GSM8K and MATH datasets, as shown in Table \ref{table:result_on_origin}. 
\textcolor{black}{Our dataset achieved the best overall performance, most frequently ranking in the top two.}
Due to limited instruction-following ability, the Mistral-7B and Llama-3-8B baselines perform poorly, but show significant improvement after fine-tuning.
% Notably, Llama-3-70b shows a 68\% improvement on GSM8K and 23\% improvement on MATH when trained on our dataset, demonstrating the effectiveness of our data. 
% However, relative to most existing works that use larger datasets, the smaller dataset may have caused Qwen-2-7b to overfit, leading to a decrease in performance on MATH. 
\textcolor{black}{We observe that fine-tuning Qwen-2-7B on most datasets results in performance degradation, which is likely due to catastrophic forgetting. In Appendix \ref{appen:cata_forget}, we provide a detailed analysis of this issue.}

\begin{figure}[!t]
    \centering
    % \begin{subfigure}[t]{0.23\textwidth}
    %     \centering
    %     \includegraphics[width=1.1\linewidth]{figures/Plot_steps_llama_3_8b_GSM8K}
    % \end{subfigure}
    % \hfill  
    \begin{subfigure}[t]{0.22\textwidth}
        \centering
        \includegraphics[width=1.1\linewidth]{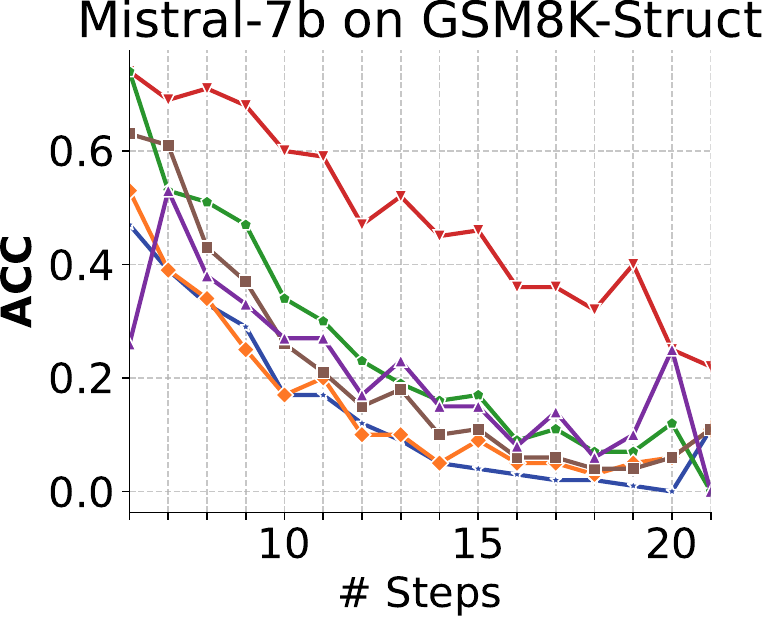}
    \end{subfigure}
    \hfill  
    \begin{subfigure}[t]{0.22\textwidth}
        \centering
        \includegraphics[width=1.1\linewidth]{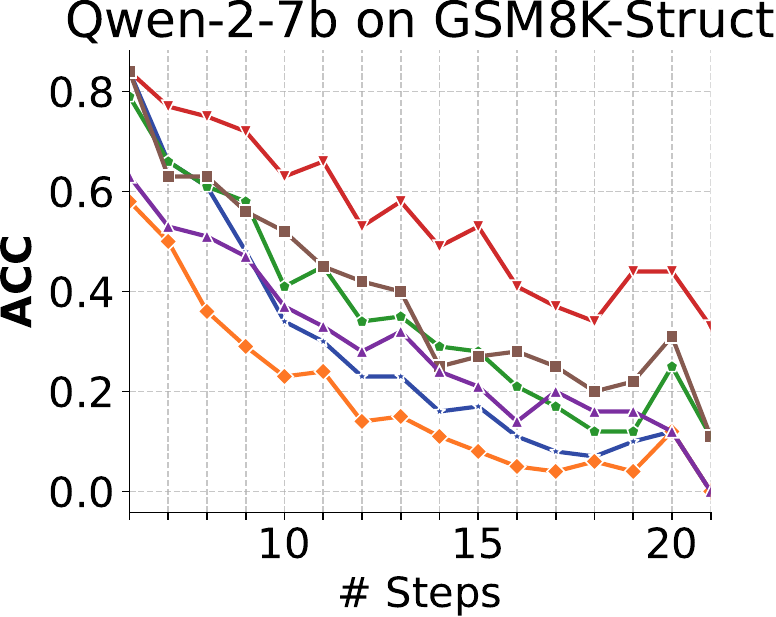}
    \end{subfigure}
    % \hfill  
    % \begin{subfigure}[t]{0.23\textwidth}
    %     \centering
    %     \includegraphics[width=1.1\linewidth]{figures/Plot_steps_llama_3_70b_GSM8K}
    % \end{subfigure}
    % \begin{subfigure}[t]{0.23\textwidth}
    %     \centering
    %     \includegraphics[width=1.1\linewidth]{figures/Plot_steps_llama_3_8b_MATH}
    % \end{subfigure}
    % \hfill  
    \begin{subfigure}[t]{0.22\textwidth}
        \centering
        \includegraphics[width=1.1\linewidth]{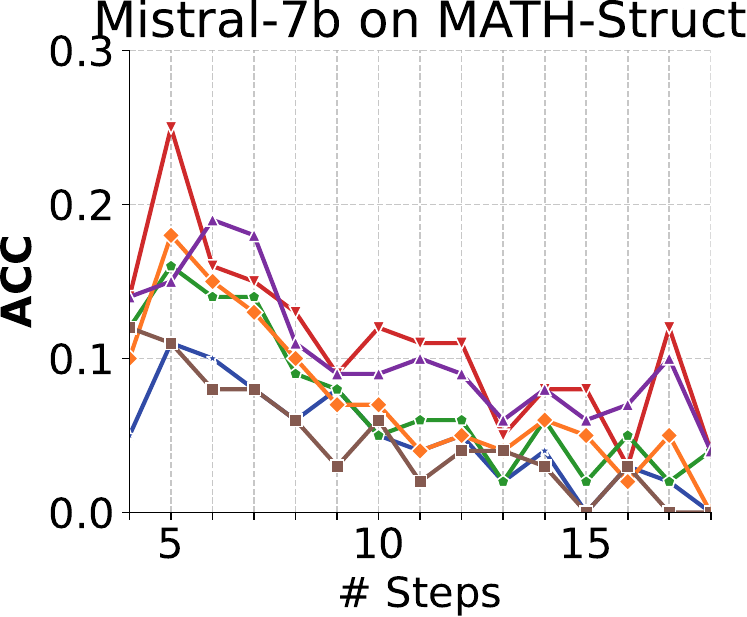}
    \end{subfigure}
    \hfill  
    \begin{subfigure}[t]{0.22\textwidth}
        \centering
        \includegraphics[width=1.1\linewidth]{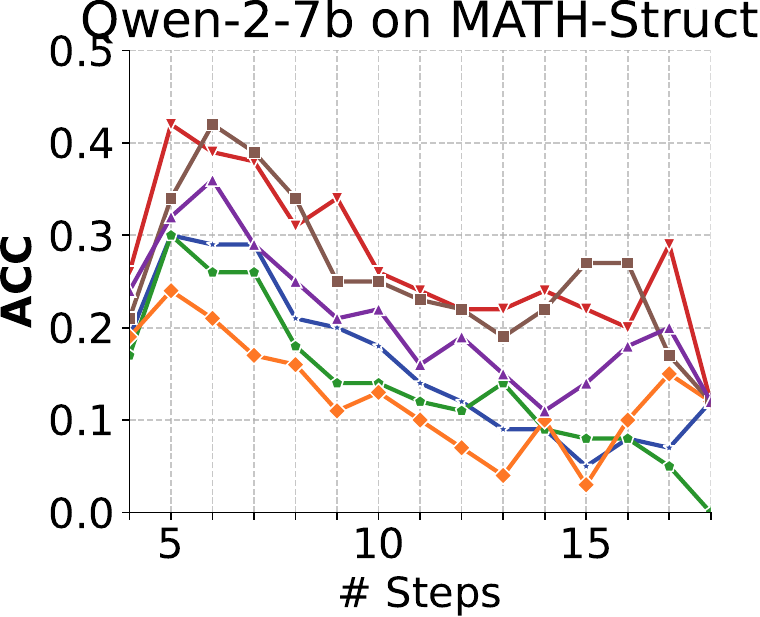}
    \end{subfigure}
    \begin{subfigure}[t]{0.4\textwidth}
        \centering
        \includegraphics[width=1.1\linewidth]{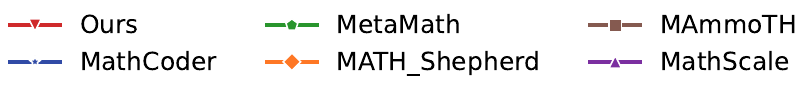}
    \end{subfigure}
    % \hfill  
    % \begin{subfigure}[t]{0.23\textwidth}
    %     \centering
    %     \includegraphics[width=1.1\linewidth]{figures/Plot_steps_llama_3_70b_MATH}   
    % \end{subfigure}
    \caption{Results of finetuned models on the proposed benchmark.}
    \label{fig:result_on_different_datasets}
\end{figure}
\subsection{Results on Different Steps}
We further evaluate the performance of the finetuned models on the proposed benchmark and present the performance distribution of each model across problems of varying difficulty in Figure \ref{fig:result_on_different_datasets}. Similar to Figure~\ref{fig:results_steps}, all methods show a decline in performance as the reasoning length of the problems increases. However, overall, the models finetuned with our data perform better across problems of all difficulty levels. Notably, for the Mistral-7b and Qwen-2-7b models, MetaMath, MAmmoTH, and our approach demonstrate comparable reasoning performance in easier problems, but their performance quickly degrades as the problem difficulty decreases. This suggests that our data helps improve LLMs' long-range reasoning capabilities.

\subsection{Versatility of the Framework}
\textcolor{black}{
Given the widespread use of GSM8K and MATH in data generation research and contain a large amount of QA data with step-by-step reasoning processes, we base our data generation on these two datasets. However, we emphasize that our method is not limited to GSM8K and MATH. To demonstrate its broader applicability, we applied our generation framework to harder math problems, and present a generated example on the AIME\_2024  \cite{AIME} dataset below. Due to space constraints, we only show the differences between the original and the generated problems.}
\begin{lstlisting}
Origin: ...She wins a prize if at least two of her numbers were $2$ of the randomly chosen numbers, and wins the grand prize if all four of her numbers were the randomly chosen numbers...
\end{lstlisting}
\vspace{10pt}
\begin{lstlisting}
New: ...She wins a small prize if exactly two of her numbers were among the randomly chosen numbers, a medium prize if exactly three numbers match, and the grand prize if all four of her numbers match...
\end{lstlisting}
\textcolor{black}{
Furthermore, our method extends beyond mathematical problems and can be applied to any task where the reasoning process can be represented through code. To illustrate this versatility, we show a programming problem generated by our intervention framework based on the MBPP \cite{MBPP} dataset:}
\begin{lstlisting}
Origin: Write a function to find minimum of three numbers.

New: Write a function that finds the minimum and maximum of three numbers and returns a tuple containing: (1) the minimum value, (2) the count of how many times this minimum value appears, (3) the maximum value, and (4) the count of how many times this maximum value appears among the three numbers. Additionally, if the minimum and maximum are the same number (meaning all three numbers are equal), return (-1, -1, -1, -1).
\end{lstlisting}
\textcolor{black}{
For the full questions and solutions of the above examples, please refer to Appendix \ref{appen:harder_example}.}

\section{Concluding Remarks}
In this paper, we explore the role of structural information in mathematical reasoning and demonstrate its various applications and importance. Based on structural data, we can quantify problem difficulty by counting reasoning steps, obtain supervision at intermediate steps, and intervene in the structure to generate more challenging samples. To the best of our knowledge, no existing work has quantitatively assessed and intervened in the difficulty of reasoning problems using structural information. We emphasize that the utility of structural information extends beyond these applications. By using code as a medium, the reasoning structure forms a narrower and more predictable channel than natural language, allowing us to trace how current variables affect subsequent computations.
% Note that the Gao et al. have already made preliminary use of this property in their PAL method  \cite{PAL}, where they altered the values of variables in the problem and obtained the intervened answers through code. 
Additionally, the inherent directed acyclic nature of computational structures provides a topological order over reasoning steps, which can be used to guide and constrain structural interventions. We plan to further explore the potential of this topological structure in future work.
% \begin{table}[!t]
%  \centering
%     \resizebox{\columnwidth}{!}{
% \begin{tabular}{cccccc}
% \toprule
%                      Model        & Dataset & Baseline   & MetaMath   & MathCoder  & Ours       \\ \midrule
% \multirow{2}{*}{Mistral-7b}  & GSM8K   & 8.72\%     & \underline{60.65\%}    & 40.26\%    & \textbf{68.31\%}    \\
%                              & MATH    & 4.50\%     & \underline{16.46\%}    & 12.30\%    & \textbf{16.90\%}    \\ \midrule
% \multirow{2}{*}{Qwen-2-7b}   & GSM8K   & 67.55\%    & \underline{69.83\%}    & 61.87\%    & \textbf{78.01\%}    \\
%                              & MATH    & \textbf{46.20\% }   & 27.66\%    & 31.16\%    & \underline{39.70\%}    \\ \midrule
% \multirow{2}{*}{Llama-3-8b}  & GSM8K   & 7.13\%     & \textbf{72.18\%}    & 46.25\%    & \underline{71.34\%}    \\
%                              & MATH    & 4.66\%     & \textbf{23.68\%}    & 16.94\%    &\underline{19.70\%}    \\ \midrule
% \multirow{2}{*}{Llama-3-70b} & GSM8K   & 20.09\% & 35.18\% & \underline{68.92\%} & \textbf{88.55\%} \\
%                              & MATH    & 14.90\% &18.44\% & \underline{33.68\%} & \textbf{38.62\%} \\ \bottomrule
% \end{tabular}}
% 	\caption{Performance on original GSM8K and MATH datasets, where the best results are in \textbf{bold} and second best are \underline{underlined}.}
% \label{table:result_on_origin}
% \end{table}

% \section{Conclusion}
% % 

\clearpage

\section*{Limitations}

Compared to other data generation approaches, the main limitation of our study is the low efficiency of the generation process, largely due to the multi-level validation strategy used to ensure data quality. Since the generated data may contain instances where individuals are assigned an excessive workload or excessive amounts of food, which, while not affecting the arithmetic itself, are often flagged as incorrect by GPT-4o based on commonsense judgment. Consequently, only 51.4\% of GSM8K and 35.4\% of MATH samples are retained. Despite being successfully generated, around 42\% of the MATH generated results and 21\% of the GSM8K generated results are discarded after GPT-4o evaluation. To ensure data quality, we did not use these discarded samples in our analysis. However, it is worth noting that since our method potentially ensures the correctness of the calculation process, these discarded data still contain supervised processes with correct calculations. As pointed out in  \cite{METAMATH}, erroneous reasoning processes may still help improve the model’s reasoning ability. \textcolor{black}{To verify this, we combine failed and successful generations to train LLMs on the mixed dataset. Results and analysis are provided in Appendix~\ref{appen:train_with_failure}. The results show that the inclusion of erroneous data appears to improve performance on MATH problems, which suggests potential for further exploration in future work.}

Another factor contributing to the low generation efficiency is that our process is incremental rather than repetitive: the $(k+1)$-th generation step builds on samples from the $k$-th intervention, progressively increasing problem difficulty. Although our generation process mitigates the model's reasoning capabilities, as the number of reasoning steps increases, the model gradually struggles to understand the presented problem and reasoning steps, making the generation process more prone to failure. While repeatedly intervening may help increase the data volume, as we mentioned before, this leads to a significant amount of discarded data. Considering that our generated data involves reasoning processes distinct from the original problems and at higher difficulty levels, a better approach might be to combine it with existing methods based on paraphrasing, thus more efficiently increasing the data volume. This will be a direction for our future research.

% \section*{Acknowledgments}

% Bibliography entries for the entire Anthology, followed by custom entries
% \bibliography{anthology,custom}
% Custom bibliography entries only
\bibliography{custom}

\clearpage
\appendix
\begin{figure*}[!htbp]
    \centering
    \includegraphics[width=\linewidth]{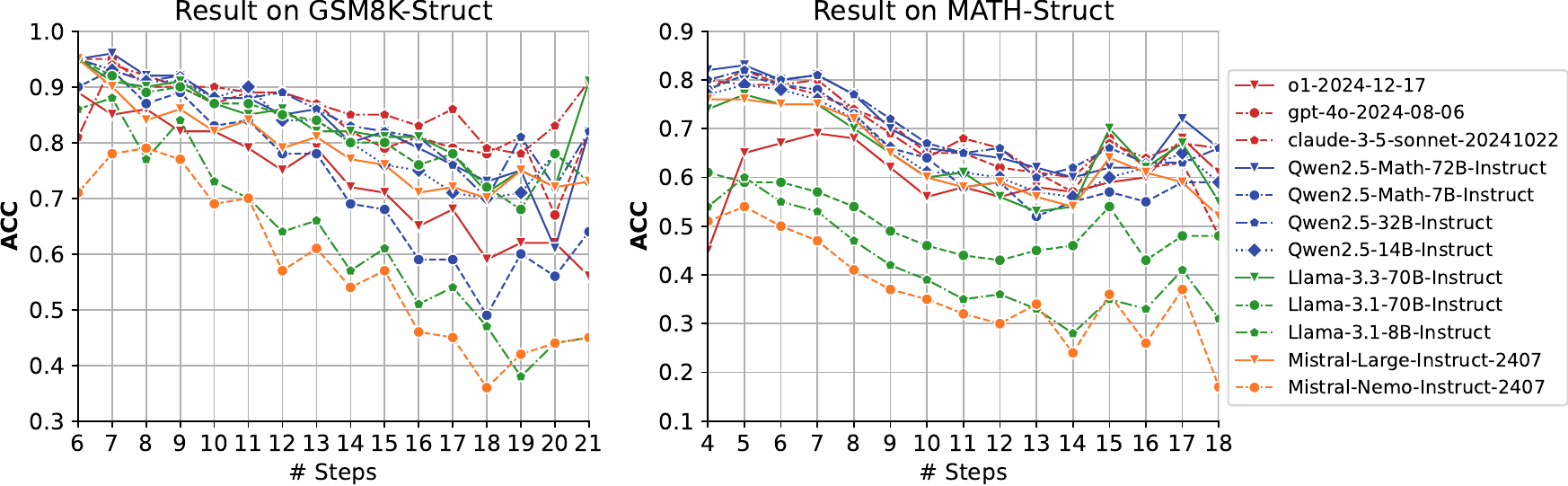}
    \caption{Additional benchmarking results.}
    \label{fig:results_additional_benchmarking}
\end{figure*}
\begin{figure*}[!htbp]
    \centering
    \includegraphics[width=\linewidth]{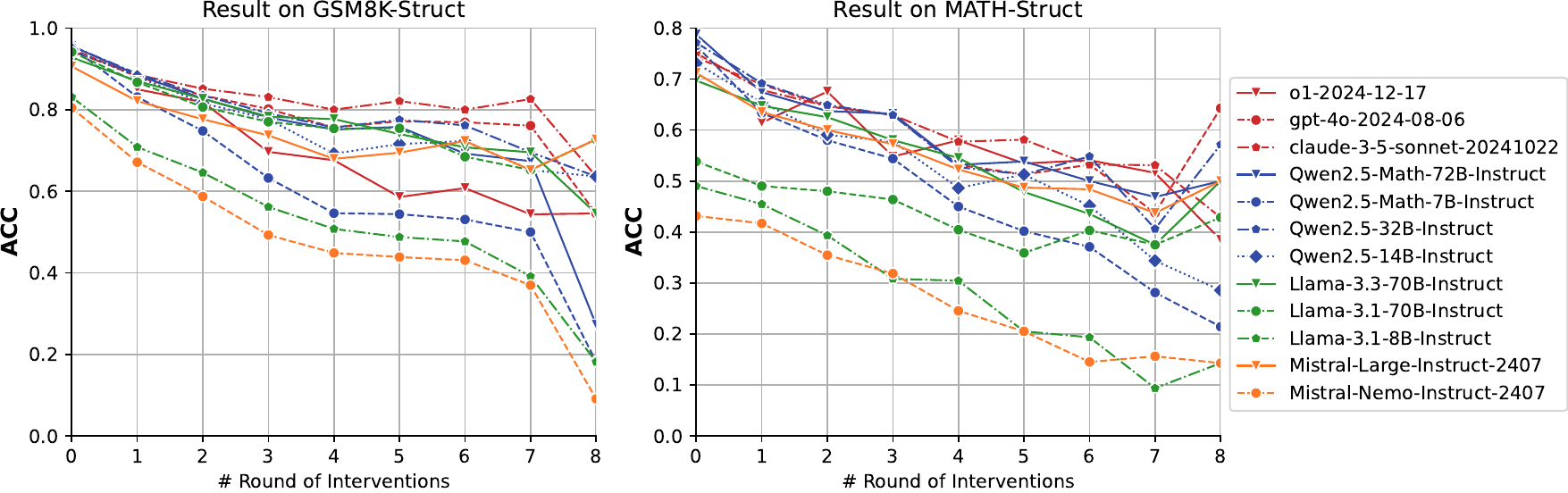}
    \caption{Benchmarking results on different number of interventions.}
    \label{fig:results_num_interventions}
\end{figure*}
\begin{figure*}[!htbp]
    \centering
    \includegraphics[width=\linewidth]{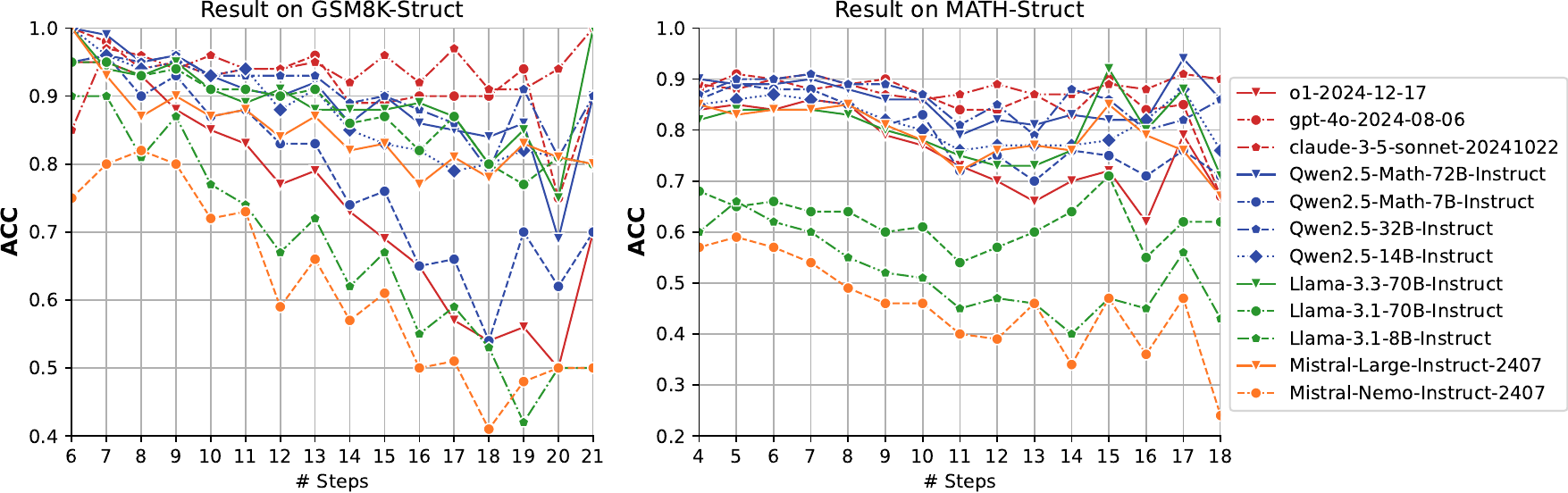}
    \caption{Benchmarking results of filtered subset on different reasoning steps.}
    \label{fig:results_bounded}
\end{figure*}

\section{Additional Benchmarking Results}
\subsection{Result on Additional Models}\label{appen:add_benchmark_results}
\textcolor{black}{In this section, we evaluate a broader range of models using our dataset,} including models from the GPT series (\verb|GPT-o1|, \verb|GPT-4o|)   \cite{GPTO1, GPT4O}, the Claude series (\verb|Claude-3-5-Sonnet|)  \cite{CLAUDE35}, the Qwen series (\verb|Qwen2.5-MATH-Instruct|, \verb|Qwen-2.5-Instruct|)  \cite{QWEN25}, the Llama-3 series (\verb|Llama-3.3|, \verb|Llama-3.1|  \cite{grattafiori2024llama3herdmodels}), and the Mistral series (\verb|Mistral-Large-Instruct|, \verb|Mistral-Nemo-Instruct|)  \cite{MISTRALLARGE, MISTRALNEMO}. 
The evaluation results are shown in Figure \ref{fig:results_additional_benchmarking}.
Comparing models within the same series, we observe that larger models generally perform better on problems requiring longer reasoning steps. This suggests that models with more parameters possess stronger long-step reasoning capabilities.
% 为了更加清晰地展示这个趋势，我们在图3中仅选取了具有代表性的模型。我们在附录X中提供了更多模型的测试结果。

\subsection{Result on Different Number of Interventions}
\label{sec:res_on_interv_num}
Recall that we generate data using a progressive approach, where each subsequent intervention is based on the data generated by the previous one. In this section, we present the performance of different models on the intervention data at each stage. As shown in Figure~\ref{fig:results_num_interventions}, the model performance generally decreases as the intervention stage advances, indicating that the difficulty of the generated data increases with each stage.

\subsection{Result on Filtered Subset}
\label{sec:res_on_filtered}
Considering that errors may still occur during the model generation process, we further validate the reliability of the evaluation results by filtering the data based on the performance of Claude-3-5-Sonnet and GPT-4o. Specifically, we assume that questions correctly answered by Claude-3-5-Sonnet and GPT-4o are correct. Based on this assumption, we select the questions correctly answered by Claude-3-5-Sonnet and GPT-4o, and take the union of both as the filtered correct data subset. We then observe the performance changes of all models on this subset, as shown in Figure~\ref{fig:results_bounded}. Compared to Figure~\ref{fig:results_steps}, we can see that the performance trend on the filtered dataset is similar to that on the full dataset. This further suggests that the increase in the number of reasoning steps raises the difficulty of the questions, leading to a gradual decline in model performance.
\section{Quality Analysis}
\label{appen:quality_analysis}
\subsection{Human Evaluation}
To further validate the quality of our dataset, we conduct a human evaluation. We randomly sampled 30 instances from both MATH-Struct and GSM8K-Struct. To ensure the quality of the evaluation, we invited two volunteer annotators to assess the dataset, specifically:
(1) Qualified annotators: Both annotators are highly qualified, with over four years of experience in computer science.
(2) Inter-annotator agreement: Annotators collaborated to resolve discrepancies and ensure consistency, maintaining high agreement and reliability.

For each question, we provide the corresponding ground truth answer and reasoning process. Annotators must then determine whether the question:
\begin{enumerate}
    \item Valid question (VQ): The problem is clear, unambiguous, and can be reasonably solved.  
    \item Unclear or ambiguous question (UAQ): The problem is vague or open to multiple interpretations.  
    \item Incomplete question (IQ): The problem lacks key information, making it unsolvable.  
    \item Multiple possible answers (MPA): The problem allows multiple correct solutions due to missing constraints.  
    \item Mismatch with ground truth (MGT): The problem does not align with the given ground truth, making evaluation unreliable.
\end{enumerate}

\begin{figure}[t]
\centering
  \includegraphics[width=.75\columnwidth]{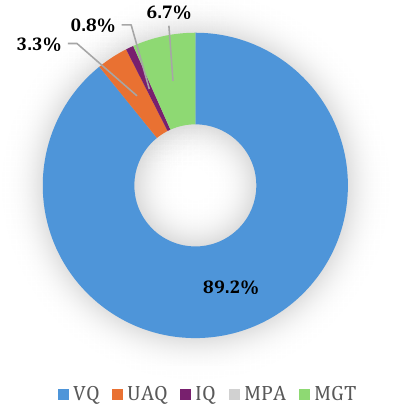}
  \caption{Human evaluation result.
    }
    \label{fig_app:human}
\end{figure}

The result of our human evaluation is presented in Figure \ref{fig_app:human}. It can be observed that valid questions (VQ) constitute nearly 90.0\%, further confirming the high quality of our dataset.

\subsection{Examples of Harder Problems}
From Figure~\ref{fig:appen_gsm8K_gen} to Figure~\ref{fig:appen_math_gpt}, we present examples where Claude-3-5-Sonnet and GPT-4o fail to answer correctly, which demonstrates that our approach can generate questions that are beyond agent’s capabilities.

\subsection{Generation on Harder Problems and Coding Problems}\label{appen:harder_example}
\textcolor{black}{
Given the widespread use of GSM8K and MATH in data generation research and contain a large amount of QA data with step-by-step reasoning processes, we base our data generation on these two datasets. However, our method is not limited to GSM8K and MATH.} To demonstrate its broader applicability, we applied our framework to harder mathematical problems (\verb|AIME_2024| \cite{AIME}) and present a generated example in Figure~\ref{fig:appen_aime}. In the generated question, the added step \verb|istotal_medium_or_better = ways_three_mat| \verb|ch + ways_four_match| modifies the conditional probability calculation to consider only cases with at least a medium prize (3 or 4 matches), unlike the original problem which includes all prize-winning cases (2, 3, or 4 matches).

\textcolor{black}{
Moreover, our method is applicable to any task where the reasoning process can be represented through code. To demonstrate this versatility, we applied it to programming problems. Figure~\ref{fig:appen_mbpp} presents the outputs of our framework on the MBPP \cite{MBPP} dataset.}

\subsection{Qualitative Error Analysis of GPT-o1}
\label{appen:analysis_of_o1}

Figure \ref{app:o1_error_gsm8k_1} to \ref{app:o1_error_math_2} include some typical cases of incorrect answers given by OpenAI's GPT-o1 model.

Figure \ref{app:o1_error_gsm8k_1} reveals that the model fails to account for the phrase ``but changed her mind and decided not to buy any utensils'', leading to an incorrect total price calculation. This indicates a weakness in understanding negations and contextual changes, suggesting that the model struggles with processing modifications to prior information. Improving its ability to recognize and adjust for such contextual shifts would enhance accuracy in similar tasks.

The issue from Figure \ref{app:o1_error_gsm8k_2} to Figure \ref{app:o1_error_math_2} highlights a key weakness: the model struggles with large numerical calculations, resulting in errors. This suggests a limitation in its ability to handle complex arithmetic accurately.

\section{Prompts}
\subsection{Prompt for Code Generation}
\label{appen:code_gen}
We present the prompt of code generation in Figure~\ref{fig:prompt_code_gen}.

\subsection{Prompt for Code Translation}
\label{appen:code_trans}
% \subsection{Guide Agent with Commented Groundtruth}
% \label{appen:commented_gt}
% - 对于Replacing Value和Reasoning Insertion，为了让模型不强行凑答案，我们在中间代码后面的注释中加入代码运行结果，并刻意不提供最终的结果
%  - 中间结果用于引导模型进行正确的推理和计算
%  - 不提供最终结果用于验证推理过程的生成是否正确。

% \subsection{Prompt for Code Translation}
We present the prompt of code generation for GSM8K dataset in Figure~\ref{fig:prompt_code_trans_GSM8K} and the prompt of code generation for MATH dataset in Figure~\ref{fig:prompt_code_trans_MATH}. Due to the higher specialization and complexity of problems in MATH, we chose to let the agent determine the specific interventions, so as to generate more reasonable problems.

% \section{Proxy Variables in Structural Intervention}
% \label{appen:proxy}

\subsection{Prompt for GPT Evaluation}
\label{appen:prompt_for_evaluation}

We provide the prompt of evaluation for GPT-4o in Figure~\ref{fig:prompt_gpt_eval}.

\subsection{Prompt for Training and Inferencing}
\label{appen:prompt_for_training}

We use the same prompt format for training and inference, which is provided in Figure~\ref{fig:prompt_train}.

% \section{GPT vs. Claude}
% \label{appen:agent_diff}
% %- GPT的干预数据质量没有Claude好，到时候可以加个附录
% %  - 增加的步骤太简单，或是直接将数字替换（见CHECK_MATH_multi_step.ipynb）
% %  - 情节描述的质量也不如Claude（见GSM8K数据 instead of）

\section{Analysis on Catastrophic Forgetting}\label{appen:cata_forget}
\textcolor{black}{
In Table~\ref{table:result_on_origin}, we observe a performance drop when fine-tuning Qwen-2-7B on most datasets, which we attribute to catastrophic forgetting. To validate this, we measured the proportion of MATH and GSM8K samples that were answered correctly by the pretrained model but incorrectly after fine-tuning with our dataset. As shown in Table~\ref{tab:flip}, about 9.1\% of MATH samples flipped from correct to incorrect. Notably, the pretrained model achieves 46.2\% accuracy on MATH—6.5\% higher than the fine-tuned model (39.7\%)—which is less than the observed forgetting rate. This supports our hypothesis that the performance loss is primarily due to knowledge forgetting during fine-tuning.}

\begin{table}[!htbp]
    \centering
    \resizebox{0.4\columnwidth}{!}{
    \begin{tabular}{cc}
    \toprule
        GAM8K & MATH\\
        \midrule
        2.5\% & 9.1\%\\
        \bottomrule
    \end{tabular}
    }
    \caption{Proportion of misled answers in MATH and GSM8K datasets after finetuning.}
    \label{tab:flip}
\end{table}

\textcolor{black}{
To further support this conclusion, we present outputs from both the pretrained and fine-tuned Qwen-2-7B models on the same question. As shown in Figure~\ref{fig:appen_cata_forget}, the pretrained model tends to produce longer responses and actively draws on its background knowledge to solve the problem. In contrast, the fine-tuned model provides a more constrained answer focused solely on the immediate problem. Given that solving MATH problems often requires retrieving and applying mathematical knowledge, we hypothesize that fine-tuning suppresses this retrieval ability, contributing to the observed performance drop.}

\section{Training with Failure Generations}
\label{appen:train_with_failure}
 \textcolor{black}{To explore the potential value of erroneous data, we combined failed and successful generations to train LLMs and evaluated their performance on the MATH and GSM8K datasets. As shown in Table \ref{tab:finetune_with_error}, while the combined data slightly decreased performance on GSM8K, it led to an improvement on MATH, suggesting that erroneous data may still contribute useful learning signals in more complex reasoning tasks. These results suggest that introducing incorrect data can further reinforce specialized mathematical knowledge in MATH, leading to improved model performance. } 

\begin{table}[!htbp]
\resizebox{\columnwidth}{!}{
\begin{tabular}{cccc}
\hline
                            & Dataset & Success (39K) & Success+Failure (100k) \\ \hline
\multirow{2}{*}{Mistral-7b} & GSM8K   & 68.31\%       & 67.70\% $\downarrow$               \\
                            & MATH    & 16.90\%       & 18.62\% $\uparrow$               \\ \hline
\multirow{2}{*}{Llama-3-8b} & GSM8K   & 71.34\%       & 70.51\% $\downarrow$               \\
                            & MATH    & 19.70\%       & 22.66\% $\uparrow$               \\ \hline
\end{tabular}
}
\caption{Finetuning with failure generations.}
\label{tab:finetune_with_error}
\end{table}

\clearpage
\begin{figure*}[t]
\begin{tcolorbox}[colback=blue!12!white,colframe=blue!75!black,title=Example from GSM8K]
\vspace{-1mm}
\begin{small}
\textbf{Original Question:}

A robe takes 2 bolts of blue fiber and half that much white fiber.  How many bolts in total does it take?

\textbf{Answer:}

\begin{tcolorbox}[colback=black!5!white,colframe=black!75!black,title=Original Solution Code]
def solution():

\quad \# <reason>The robe takes 2 bolts of blue fiber.</reason>

\quad blue\_fiber\_bolts = 2 \# 2

\quad \# <reason>The white fiber needed is half of the blue fiber, so it takes 2/2=<<2/2=1>>1 bolt of white fiber.</reason>

\quad white\_fiber\_bolts = blue\_fiber\_bolts / 2 \# 1.0

\quad \# <reason>The total amount of fabric needed is 2+1=<<2+1=3>>3 bolts.</reason>

\quad total\_bolts = blue\_fiber\_bolts + white\_fiber\_bolts \# 3.0

\quad \# <reason>The result is 3 bolts in total.</reason>

\quad result = total\_bolts \# 3.0

\quad return result
\end{tcolorbox}

\textbf{New Question:}

A grand ceremonial robe takes 2 bolts of blue fiber. The design first needs to be made 5 times larger. Due to a rare celestial event, the design must be made 9 times more intricate before being made 6 times more elaborate for the winter festival. The white fiber needed is still half of the blue fiber amount. Due to an unexpected surge in festival attendance, the grand master needs 162 extra bolts. Furthermore, due to a special royal decree, an additional 810 bolts are required for ceremonial banners. How many more bolts in total are needed compared to the original estimate of 18 bolts?

\textbf{Answer:}
\begin{tcolorbox}[colback=black!5!white,colframe=black!75!black,title=New Solution Code]
def solution():

\quad \# <reason>The basic robe design starts with 2 bolts of blue fiber.</reason>

\quad blue\_fiber\_bolts = 2  \# 2

\quad \# <reason>First, the design is made 5 times larger, requiring 2 * 5 = 10 bolts.</reason>

\quad larger\_design\_bolts = blue\_fiber\_bolts * 5  \# 10

\quad \# <reason>Due to the celestial event, the design must be made 9 times more intricate.</reason>

\quad op\_var = 9  \# 9

\quad \# <reason>The intricate design now requires 10 * 9 = 90 bolts.</reason>

\quad extra\_var = larger\_design\_bolts * op\_var  \# 90

\quad \# <reason>Then, for the festival version, we need 6 times as much blue fiber as the intricate design, so 90 * 6 = 540 bolts of blue fiber.</reason>

\quad festival\_blue\_fiber\_bolts = extra\_var * 6  \# 540

\quad \# <reason>The white fiber needed is half of the blue fiber, so it takes 540/2 = 270 bolts of white fiber.</reason>

\quad white\_fiber\_bolts = festival\_blue\_fiber\_bolts / 2  \# 270.0

\quad \# <reason>The new design requires a total amount of 540 + 270 = 810 bolts.</reason>

\quad total\_bolts = festival\_blue\_fiber\_bolts + white\_fiber\_bolts  \# 810.0

\quad \# <reason>The initial difference between the new design requirement (810 bolts) and the original estimate (18 bolts) is 810 - 18 = 792 bolts.</reason>

\quad bolts\_difference = total\_bolts - 18.0  \# 792.0

\quad \# <reason>With the 162 extra bolts needed for increased attendance, the total additional requirement becomes 792 + 162 = 954 bolts more than the original estimate.</reason>

\quad additional\_bolts\_needed = bolts\_difference + 162.0  \# 954.0

\quad \# <reason>Adding the 810 bolts required by the royal decree for ceremonial banners, we need 954 + 810 = 1764 more bolts than originally estimated.</reason>

\quad total\_additional\_bolts = additional\_bolts\_needed + 810.0  \# ?

\quad \# <reason>The final result is 1764 bolts more than originally estimated.</reason>

\quad result = total\_additional\_bolts  \# ?

\quad return result

\end{tcolorbox}

\end{small}
% \vspace{-1mm}
\end{tcolorbox}
\caption{Extended question from GSM8K.}
\label{fig:appen_gsm8K_gen}
\end{figure*}

\begin{figure*}[t]
\begin{tcolorbox}[colback=blue!12!white,colframe=blue!75!black,title=Example from GSM8K]
\vspace{-1mm}
\begin{tcolorbox}[colback=red!5!white,colframe=red!75!black,title=Answer by Claude-3-5-Sonnet]
\vspace{-1mm}
\begin{small}
\textbf{Answer:}

Let me help solve this step by step.

1) Let's first establish the base amount:

\quad - Original estimate = 18 bolts

2) Calculate how the design changes affect the amount:

\quad - 5 times larger × 9 times more intricate × 6 times more elaborate

\quad - 5 × 9 × 6 = 270 times the original amount

3) Calculate new amount after design changes:

\quad - 18 × 270 = 4,860 bolts

4) Add extra bolts needed:

\quad - Additional due to surge: 162 bolts

\quad - Additional for banners: 810 bolts

\quad - Total extra: 162 + 810 = 972 bolts

5) Calculate total new amount:

\quad - 4,860 + 972 = 5,832 bolts

6) Find difference from original estimate:

\quad - 5,832 - 18 = 5,814

Therefore, 5,814 more bolts are needed compared to the original estimate.
\LARGE \textcolor{red}{\XSolidBrush}

\end{small}
% \vspace{-1mm}
\end{tcolorbox}

\begin{tcolorbox}[colback=red!5!white,colframe=red!75!black,title=Answer by GPT-4o]
\vspace{-1mm}
\begin{small}
\textbf{Answer:}

To solve the problem step by step, let's break down each part of the problem and make calculations as necessary:

1. Initial Blue Fiber Requirement:

\quad - Initial requirement for the robe is 2 bolts of blue fiber.

2. Design Enlargement:

\quad - The design needs to be made 5 times larger. If the base requirement is 2 bolts, then $2 \times 5 = 10$ bolts are needed just for the size increase.

3. Increased Intricacy:

\quad - The design must be made 9 times more intricate. Hence, multiplying the enlarged design's requirement by 9 gives us $10 \times 9 = 90$ bolts.

4. Elaboration for the Winter Festival:

\quad - The design must then be made 6 times more elaborate. Therefore, multiplying the requirement by 6 yields: $90 \times 6 = 540$ bolts of blue fiber.

5. White Fiber Requirement:

\quad - The white fiber required is half of the blue fiber amount. Therefore, if the blue fiber needed is 540 bolts, then the white fiber requirements would be $\frac{540}{2} = 270$ bolts.

6. Total Fiber Requirement (excluding extras):

\quad - Blue fiber: 540 bolts

\quad - White fiber: 270 bolts

\quad - Total without extras: $540 + 270 = 810$ bolts

7. Additional Bolt Requirements:

\quad - An additional 162 bolts are needed due to a surge in festival attendance.

\quad - An additional 810 bolts are required due to a special royal decree for ceremonial banners.

8. Total Additional Bolts:

\quad - Additional from festival attendance: 162 bolts

\quad - Additional from royal decree: 810 bolts

\quad - Total additional bolts: $162 + 810 = 972$ bolts

9. Total Bolts Needed:

\quad - Total from the enlarged, intricate, and elaborate design plus additional requirements: $810 + 972 = 1782$ bolts

10. Comparison with the Original Estimate:

\quad - The original estimate is 18 bolts.

11. Difference in Requirement:

\quad - Total more bolts needed compared to the original estimate: $1782 - 18 = 1764$.

Therefore, the total number of additional bolts needed compared to the initial estimate is \verb|\|boxed\{1764\}. \LARGE \textcolor{teal}{\Checkmark}

\end{small}
% \vspace{-1mm}
\end{tcolorbox}

% \vspace{-1mm}
\end{tcolorbox}
\caption{Generated answers by Claude-3-5 Sonnet and GPT-4o.}
\label{fig:appen_gsm8K_ans}
\end{figure*}

\begin{figure*}[t]
\begin{tcolorbox}[colback=blue!12!white,colframe=blue!75!black,title=Example from MATH]
\vspace{-1mm}
\begin{small}
\textbf{Original Question:}

If $5x - 3 = 12$, what is the value of $5x + 3$?

\textbf{Answer:}

\begin{tcolorbox}[colback=black!5!white,colframe=black!75!black,title=Original Solution Code]
def solution():

\quad \# <reason>From the equation $5x - 3 = 12$, we can solve for $5x$ by adding 3 to both sides.</reason>

\quad right\_hand\_side = 12

\quad constant\_term = 3

\quad \# <reason>Adding 3 to both sides gives $5x = 12 + 3 = 15$.</reason>

\quad five\_x = right\_hand\_side + constant\_term

\quad \# <reason>Now for the expression $5x + 3$, we take the value of $5x$ which is 15 and add 3.</reason>

\quad \# <reason>Therefore, $5x + 3 = 15 + 3 = $\verb|\|boxed\{18\}.</reason>

\quad result = five\_x + constant\_term

\quad return result
\end{tcolorbox}

\textbf{New Question:}

If $5x - 3 = 12$, what is the value of $(5x + 3)^5 - (5x - 3)^5 + (5x)^5$?

\textbf{Answer:}
\begin{tcolorbox}[colback=black!5!white,colframe=black!75!black,title=New Solution Code]

def solution():

\quad \# <reason>From the equation $5x - 3 = 12$, we can solve for $5x$ by adding 3 to both sides.</reason>

\quad right\_hand\_side = 12

\quad constant\_term = 3

\quad \# <reason>Adding 3 to both sides gives $5x = 12 + 3 = 15$.</reason>

\quad five\_x = right\_hand\_side + constant\_term

\quad \# <reason>Now for the first expression $5x + 3$, we take the value of $5x$ which is 15 and add 3.</reason>

\quad expression1 = five\_x + constant\_term

\quad \# <reason>Therefore, $5x + 3 = 15 + 3 = 18$.</reason>

\quad \# <reason>For the second expression $5x - 3$, we already know it equals 12.</reason>

\quad expression2 = right\_hand\_side

\quad \# <reason>For the third expression $5x$, we know it equals 15.</reason>

\quad expression3 = five\_x

\quad \# <reason>For fifth powers, we can use the difference of fifth powers formula: $a^5 - b^5 = (a-b)(a^4 + a^3b + a^2b^2 + ab^3 + b^4)$</reason>

\quad a = expression1

\quad b = expression2

\quad \# <reason>First calculate $a-b = 18-12 = 6$</reason>

\quad difference = a - b

\quad \# <reason>Now we can compute $(5x + 3)^5 - (5x - 3)^5 = 18^5 - 12^5 = (18-12)(18^4 + 18^3*12 + 18^2*12^2 + 18*12^3 + 12^4)$</reason>

\quad result1 = difference*(a**4 + a**3*b + a**2*b**2 + a*b**3 + b**4)

\quad \# <reason>Finally, add $(5x)^5 = 15^5$ to get the final result</reason>

\quad result = result1 + expression3**5

\quad return result

\end{tcolorbox}

\end{small}
% \vspace{-1mm}
\end{tcolorbox}
\caption{Extended question from MATH.}
\label{fig:appen_math}
\end{figure*}

\begin{figure*}[t]
\begin{tcolorbox}[colback=blue!12!white,colframe=blue!75!black,title=Example from MATH]
\vspace{-1mm}
\begin{tcolorbox}[colback=red!5!white,colframe=red!75!black,title=Answer by Claude]
\vspace{-1mm}
\begin{small}
\textbf{Answer:}

Let me solve this step by step.

1) First, from $5x - 3 = 12$

\quad - Let's solve for $5x$

\quad - $5x = 15$

\quad - Therefore, $x = 3$

2) Now, let's substitute $5x = 15$ into each expression:

\quad - $5x + 3 = 18$

\quad - $5x - 3 = 12$

\quad - $5x = 15$

3) Our expression is $(5x + 3)^5 - (5x - 3)^5 + (5x)^5$

\quad - This becomes $18^5 - 12^5 + 15^5$

4) Let's calculate each term:

\quad - $18^5 = 1,889,568$

\quad - $12^5 = 248,832$

\quad - $15^5 = 759,375$

5) Now let's do the final calculation:

\quad - $1,889,568 - 248,832 + 759,375$

\quad - $1,889,568 + 759,375 - 248,832$

\quad - $2,648,943 - 248,832$

\quad - $2,400,111$

Therefore, \verb|\|boxed\{2,400,111\}  \LARGE \textcolor{teal}{\Checkmark}

\end{small}
% \vspace{-1mm}
\end{tcolorbox}
\end{tcolorbox}
\caption{Correct answer by Claude-3-5-Sonnet.}
\label{fig:appen_math_claude}
\end{figure*}

\begin{figure*}[t]
\begin{tcolorbox}[colback=blue!12!white,colframe=blue!75!black,title=Example from MATH]
\vspace{-1mm}
\begin{tcolorbox}[colback=red!5!white,colframe=red!75!black,title=Answer by GPT-4o]
\vspace{-1mm}
\begin{small}
\textbf{Answer:}

To solve the problem, we start by determining the value of $ x $ from the equation $ 5x - 3 = 12 $.

Step 1: Solve for $ x $

\[5x - 3 = 12\]

Add 3 to both sides to isolate the term with $ x $:

\[5x = 12 + 3\]
\[5x = 15\]

Divide both sides by 5 to solve for $ x $:

\[x = \frac{15}{5}\]
\[x = 3\]

Step 2: Find the values of $ 5x + 3 $, $ 5x - 3 $, and $ 5x $, and then use these to find $(5x + 3)^5 - (5x - 3)^5 + (5x)^5$

For $ x = 3 $:

\[5x = 5 \times 3 = 15\]
\[5x + 3 = 15 + 3 = 18\]
\[5x - 3 = 15 - 3 = 12\]

Now compute:

\[(5x + 3)^5 = 18^5\]
\[(5x - 3)^5 = 12^5\]
\[(5x)^5 = 15^5\]

We want to find:

\[(5x + 3)^5 - (5x - 3)^5 + (5x)^5\]

Substituting the known expressions:

\[18^5 - 12^5 + 15^5\]

Step 3: Simplify using algebraic identities

Notice that:

\[(5x + 3)^5 - (5x - 3)^5\]

can be rewritten using the identity for the difference of two powers. However, direct computation is feasible for specific integer substitutions like here.

\[= (18^5 - 12^5) + 15^5\]

we'll actually find separately $ a = 18 $, $ b = 12 $, $ c = 15 $, and verify simplified nodes. More insight into simplification is if you notice symmetric material:

Thus separate terms:

Notice there's symmetry in approximating:

If you alert yourself:

Throwback Insight: {\LARGE \textcolor{red}{\XSolidBrush} }\textcolor{red}{(Meaningless output)}

Arithmetic arithmetic:

Expertize

While these factors symmetric exposure releases:

therefore \verb|\|boxed\{0\}

\end{small}
% \vspace{-1mm}
\end{tcolorbox}
\end{tcolorbox}
\caption{Wrong answer by GPT-4o.}
\label{fig:appen_math_gpt}
\end{figure*}

\begin{figure*}[t]
\begin{tcolorbox}[colback=blue!12!white,colframe=blue!75!black,title= Example from AIME\_2024]
\vspace{-1mm}
\begin{small}
\textbf{Original Question:}

Jen enters a lottery by picking $4$ distinct numbers from $S=\{1,2,3,\cdots,9,10\}.$ $4$ numbers are randomly chosen from $S.$ She wins a prize if at least two of her numbers were $2$ of the randomly chosen numbers, and wins the grand prize if all four of her numbers were the randomly chosen numbers. The probability of her winning the grand prize given that she won a prize is $\frac{m}{n}$ where $m$ and $n$ are relatively prime positive integers. Find $m+n$.

\begin{tcolorbox}[colback=black!5!white,colframe=black!75!black,title=Original Solution Code]
def solution():

\quad \# <reason>Import math package for computing combinations.</reason>

\quad from math import comb
    
\quad \# <reason>For getting exactly 2 numbers right: We have C(4,2) ways to choose which 2 of Jen's 4 numbers match the lottery,

\quad \# and C(6,2) ways for the lottery to choose the other 2 numbers from the remaining 6 numbers.</reason>

\quad ways\_two\_match = comb(4, 2) \* comb(6, 2)
    
\quad \# <reason>For getting exactly 3 numbers right: We have C(4,3) ways to choose which 3 of Jen's 4 numbers match the lottery,
    
\quad \# and C(6,1) ways for the lottery to choose the last number from the remaining 6 numbers.</reason>

\quad ways\_three\_match = comb(4, 3) * comb(6, 1)
    
\quad \# <reason>For getting all 4 numbers right: There is only 1 way - all numbers must match.</reason>
    
\quad ways\_four\_match = 1
    
\quad \# <reason>Total number of favorable cases (winning a prize) is the sum of ways to get 2, 3, or 4 matches.</reason>
    
\quad total\_favorable\_cases = ways\_two\_match + ways\_three\_match + ways\_four\_match
    
\quad \# <reason>Using Bayes' Theorem, P(grand prize | prize) = P(grand prize) / P(prize) = 1/115.</reason>
    
\quad m = 1  \# numerator
    
\quad n = total\_favorable\_cases  \# denominator
    
\quad \# <reason>The answer is m + n = 1 + 115 = 116.</reason>
    
\quad result = m + n

\quad return result

\end{tcolorbox}

\textbf{New Question:}

Jen enters a lottery by picking $4$ distinct numbers from $S=\{1,2,3,\cdots,9,10\}.$ $4$ numbers are randomly chosen from $S.$ She wins a small prize if exactly two of her numbers were among the randomly chosen numbers, a medium prize if exactly three numbers match, and the grand prize if all four of her numbers match. The probability of her winning the grand prize given that she won at least a medium prize is $\frac{m}{n}$ where $m$ and $n$ are relatively prime positive integers. Find $m+n$.

\begin{tcolorbox}[colback=black!5!white,colframe=black!75!black,title=New Solution Code]

def solution():

\quad \# <reason>Import math package for computing combinations.</reason>
    
\quad from math import comb
    
\quad \# <reason>For getting exactly 2 numbers right: We have C(4,2) ways to choose which 2 of Jen's 4 numbers match the lottery,
    
\quad \# and C(6,2) ways for the lottery to choose the other 2 numbers from the remaining 6 numbers.</reason>
    ways\_two\_match = comb(4, 2) * comb(6, 2)

\quad \# <reason>For getting exactly 3 numbers right: We have C(4,3) ways to choose which 3 of Jen's 4 numbers match the lottery,
    
\quad \# and C(6,1) ways for the lottery to choose the last number from the remaining 6 numbers.</reason>
    
\quad ways\_three\_match = comb(4, 3) * comb(6, 1)
    
\quad \# <reason>For getting all 4 numbers right: There is only 1 way - all numbers must match.</reason>
    
\quad ways\_four\_match = 1
    
\quad \# <reason>For at least medium prize, we only consider cases with 3 or 4 matches.</reason>
    
\quad total\_medium\_or\_better = ways\_three\_match + ways\_four\_match
    
\quad \# <reason>Using Bayes' Theorem, P(grand prize | medium or better) = P(grand prize) / P(medium or better) = 1/25.</reason>
    
\quad m = 1  \# numerator
    
\quad n = total\_medium\_or\_better  \# denominator
    
\quad \# <reason>The answer is m + n = 1 + 25 = 26.</reason>
    result = m + n
    
\quad return result

\end{tcolorbox}

\end{small}
% \vspace{-1mm}
\end{tcolorbox}
\caption{Extended question from AIME\_2024.}
\label{fig:appen_aime}
\end{figure*}

\begin{figure*}[t]
\resizebox{0.98\linewidth}{!}{
\begin{tcolorbox}[colback=blue!12!white,colframe=blue!75!black,title= Example from MBPP]
\vspace{-1mm}
\begin{small}
\textbf{Original Question:}

Write a function to find minimum of three numbers.

Original Test Samples:

assert min\_of\_three(10,20,0) == 0

assert min\_of\_three(19,15,18) == 15

assert min\_of\_three(-10,-20,-30) == -30

\begin{tcolorbox}[colback=black!5!white,colframe=black!75!black,title=Original Solution Code]

def min\_of\_three(a,b,c): 
    
\quad if (a <= b) and (a <= c): 
        
\quad \quad smallest = a 
    
\quad elif (b <= a) and (b <= c): 
        
\quad \quad smallest = b 
    
\quad else: 
        
\quad \quad smallest = c 
    
\quad return smallest 
\end{tcolorbox}

\textbf{New Question:}

Write a function that finds the minimum and maximum of three numbers and returns a tuple containing: (1) the minimum value, (2) the count of how many times this minimum value appears, (3) the maximum value, and (4) the count of how many times this maximum value appears among the three numbers. Additionally, if the minimum and maximum are the same number (meaning all three numbers are equal), return (-1, -1, -1, -1).

New Test Samples:

assert min\_max\_with\_count(10, 20, 0)  ==  (0, 1, 20, 1)

assert min\_max\_with\_count(15, 15, 18)  ==  (15, 2, 18, 1)

assert min\_max\_with\_count(-30, -30, -30)  ==  (-1, -1, -1, -1)

assert min\_max\_with\_count(-20, -20, -10)  ==  (-20, 2, -10, 1)

assert min\_max\_with\_count(5, 3, 3)  ==  (3, 2, 5, 1)

assert min\_max\_with\_count(7, 7, 4)  ==  (4, 1, 7, 2)

\begin{tcolorbox}[colback=black!5!white,colframe=black!75!black,title=New Solution Code]

def min\_max\_with\_count(a, b, c):
    
\quad \# First find the minimum value
    
\quad if (a <= b) and (a <= c):
    
\quad \quad smallest = a

\quad elif (b <= a) and (b <= c):
    
\quad \quad smallest = b
        
\quad else:
    
\quad \quad smallest = c

\quad \# Find the maximum value
    
\quad if (a >= b) and (a >= c):
        
\quad \quad largest = a
    
\quad elif (b >= a) and (b >= c):
        
\quad \quad largest = b
    
\quad else:
        
\quad \quad largest = c

\quad \# Count occurrences of the minimum value
    
\quad min\_count = 0
    
\quad if a  ==  smallest:
        
\quad \quad min\_count += 1
    
\quad if b  ==  smallest:
        
\quad \quad min\_count += 1
    
\quad if c  ==  smallest:
        
\quad \quad min\_count += 1

\quad \# Count occurrences of the maximum value
    
\quad max\_count = 0
    
\quad if a  ==  largest:
        
\quad \quad max\_count += 1
    
\quad if b  ==  largest:
        
\quad \quad max\_count += 1
    
\quad if c  ==  largest:
        
\quad \quad max\_count += 1

\quad \# Check if all numbers are equal
    
\quad if smallest  ==  largest:
        
\quad \quad return (-1, -1, -1, -1)

\quad return (smallest, min\_count, largest, max\_count)

\end{tcolorbox}

\end{small}
% \vspace{-1mm}
\end{tcolorbox}
}
\caption{Extended question from MBPP.}
\label{fig:appen_mbpp}
\end{figure*}

\begin{figure*}[t]
\begin{tcolorbox}[colback=blue!12!white,colframe=blue!75!black,title=Error Case of GPT-o1]
\vspace{-1mm}
\begin{small}
\textbf{Question:}

Charlotte went into the kitchen supply store knowing she wanted two identical sets of pots and pans for \$120.00 each, and a set of mixing bowls for \$20.00. She initially planned to buy 5 utensils at \$5.00 each but changed her mind and decided not to buy any utensils. At check out, the clerk told her everything was 20\\% off. Two months later, Charlotte came back and spent \$709.60 more at the same store. How much did Charlotte spend in total at this store?

\textbf{Answer:}

917.6
\begin{tcolorbox}
[colback=red!5!white,colframe=red!75!black,title=O1's wrong answer]
Step 1: Calculate the initial cost of the items 

Charlotte intended to buy
Charlotte wanted:
- Two identical sets of pots and pans, each for \$120.00
- One set of mixing bowls for \$20.00
- Initially, she planned to buy 5 utensils at \$5.00 each

The cost of two sets of pots and pans:
\[
2 \times 120 = 240
\]

The cost of one set of mixing bowls:
\[
1 \times 20 = 20
\]

The cost of 5 utensils at \$5.00 each:  {\LARGE \textcolor{red}{\XSolidBrush} }\textcolor{red}{(Misunderstand question)}
\[
5 \times 5 = 25 
\]

So, initially, the total cost without any discounts is:
\[
240 + 20 + 25 = 285
\]

Step 2: Apply the 20\% discount

Since everything was 20\% off, Charlotte would only pay 80\% of the original price. To find the discounted price, we multiply the total cost by 0.80:
\[
285 \times 0.80 = 228
\]

Thus, after the discount, the total cost at checkout is:
\[
228
\]

Step 3: Calculate Charlotte's total spending after the second purchase

Two months later, Charlotte came back and spent \$709.60 more. To find her total spending, we add this to the amount she spent during her first visit:
\[
228 + 709.60 = 937.60
\]

Final Answer:
Charlotte spent a total of:
\[
\boxed{937.60}
\]

\end{tcolorbox}

\end{small}
% \vspace{-1mm}
\end{tcolorbox}
\caption{Error Case 1 of GPT-o1.}
\label{app:o1_error_gsm8k_1}
\end{figure*}

\begin{figure*}[t]
\begin{tcolorbox}[colback=blue!12!white,colframe=blue!75!black,title=Error Case of GPT-o1]
\vspace{-1mm}
\begin{small}
\textbf{Question:}

Michael is replacing the carpet in his bedroom. The carpet he initially chose costs \$12 per square foot, but due to a supply shortage, the price increased by 6 times. Additionally, he needs to pay \$2 per square foot for padding underneath. His contractor initially quoted \$4 per square foot to remove the old carpet, but due to labor shortages, this cost also increased by 6 times. The installation cost is \$34 per square foot, but due to the complexity of the job, the contractor had to triple this rate. His bedroom measures 18 feet by 12 feet. How much will it cost Michael to replace the carpet?

\textbf{Answer:}

43200
\begin{tcolorbox}
[colback=red!5!white,colframe=red!75!black,title=O1's wrong answer]
Let's break the problem into parts step by step:

Step 1: Calculate the area of the bedroom

The area of the bedroom is given by:
\[
\text{Area} = \text{Length} \times \text{Width} = 18 \, \text{ft} \times 12 \, \text{ft} = 216 \, \text{sq ft}
\]

Step 2: Calculate the cost of the carpet

Initially, the carpet costs \$12 per square foot, but due to the price increase, it is now 6 times more expensive. The new cost per square foot for the carpet is:
\[
\text{New carpet cost per square foot} = 12 \times 6 = 72 \, \text{dollars per square foot}
\]
Thus, the total cost for the carpet is:
\[
\text{Total carpet cost} = 72 \, \text{dollars per square foot} \times 216 \, \text{sq ft} = 15552 \, \text{dollars}
\]

Step 3: Calculate the cost of the padding

The padding costs \$2 per square foot, so the total cost for the padding is:
\[
\text{Total padding cost} = 2 \, \text{dollars per square foot} \times 216 \, \text{sq ft} = 432 \, \text{dollars}
\]

Step 4: Calculate the cost to remove the old carpet

Initially, the removal cost is \$4 per square foot, but due to the labor shortage, it is now 6 times more expensive. The new cost per square foot for removing the old carpet is:
\[
\text{New removal cost per square foot} = 4 \times 6 = 24 \, \text{dollars per square foot}
\]
Thus, the total removal cost is:
\[
\text{Total removal cost} = 24 \, \text{dollars per square foot} \times 216 \, \text{sq ft} = 5184 \, \text{dollars}
\]

Step 5: Calculate the installation cost

The installation cost is \$34 per square foot, but due to the complexity of the job, the rate is tripled. The new cost per square foot for installation is:
\[
\text{New installation cost per square foot} = 34 \times 3 = 102 \, \text{dollars per square foot}
\]
Thus, the total installation cost is:
\[
\text{Total installation cost} = 102 \, \text{dollars per square foot} \times 216 \, \text{sq ft} = 22032 \, \text{dollars}
\]

Step 6: Calculate the total cost

Now, we can add up all the costs:{\LARGE \textcolor{red}{\XSolidBrush} }\textcolor{red}{(Miscalculation)}
\[
\text{Total cost} = \text{Total carpet cost} + \text{Total padding cost} + \text{Total removal cost} + \text{Total installation cost}
\]
\[
\text{Total cost} = 15552 + 432 + 5184 + 22032 = 43300 \, \text{dollars}
\]

Thus, the total cost for Michael to replace the carpet is:
\[
\boxed{43300}
\]

\end{tcolorbox}

\end{small}
% \vspace{-1mm}
\end{tcolorbox}
\caption{Error Case 2 of GPT-o1.}
\label{app:o1_error_gsm8k_2}
\end{figure*}
\begin{figure*}[t]
\resizebox{0.95\linewidth}{!}{
\begin{tcolorbox}[colback=blue!12!white,colframe=blue!75!black,title=Error Case of GPT-o1]
\vspace{-1mm}
\begin{small}
\textbf{Question:}

Compute the following:
\[
\binom{16}{4} - \binom{16}{12} + \binom{16}{8} - \binom{16}{9} + \binom{16}{7}
\]

\textbf{Answer:}

12870
\begin{tcolorbox}
[colback=red!5!white,colframe=red!75!black,title=O1's wrong answer]
Step 1: Understand the binomial coefficient

The binomial coefficient \(\dbinom{n}{k}\) is defined as:

\[
\dbinom{n}{k} = \frac{n!}{k!(n-k)!}
\]

Step 2: Compute each binomial coefficient

Compute \(\dbinom{16}{4}\):
\[
\dbinom{16}{4} = \frac{16!}{4!(16-4)!} = \frac{16 \times 15 \times 14 \times 13}{4 \times 3 \times 2 \times 1} = 1820
\]

Compute \(\dbinom{16}{12}\):
By the symmetry of binomial coefficients, \(\dbinom{16}{12} = \dbinom{16}{4}\). Therefore:
$\dbinom{16}{12} = 1820$

Compute \(\dbinom{16}{8}\):
\[
\dbinom{16}{8} = \frac{16!}{8!(16-8)!} = \frac{16 \times 15 \times 14 \times 13 \times 12 \times 11 \times 10 \times 9}{8 \times 7 \times 6 \times 5 \times 4 \times 3 \times 2 \times 1} = 12870
\]

Compute \(\dbinom{16}{9}\):
\[
\dbinom{16}{9} = \frac{16!}{9!(16-9)!} = \frac{16 \times 15 \times 14 \times 13 \times 12 \times 11 \times 10 \times 9 \times 8}{9 \times 8 \times 7 \times 6 \times 5 \times 4 \times 3 \times 2 \times 1} = 11440
\]

Compute \(\dbinom{16}{7}\):
\[
\dbinom{16}{7} = \frac{16!}{7!(16-7)!} = \frac{16 \times 15 \times 14 \times 13 \times 12 \times 11 \times 10}{7 \times 6 \times 5 \times 4 \times 3 \times 2 \times 1} = 11440
\]

Step 3: Substitute the values into the original expression

Now substitute the computed values of the binomial coefficients into the expression:

\[
\dbinom{16}{4} - \dbinom{16}{12} + \dbinom{16}{8} - \dbinom{16}{9} + \dbinom{16}{7} = 1820 - 1820 + 12870 - 11440 + 11440
\]

Step 4: Simplify the expression
Now simplify the expression step by step:{\LARGE \textcolor{red}{\XSolidBrush} }\textcolor{red}{(Miscalculation)}

\[
1820 - 1820 = 0
\]
\[
0 + 12870 = 12870
\]
\[
12870 - 11440 = 1420
\]
\[
1420 + 11440 = 12860
\]

Final Answer:
The value of the expression is:
$\boxed{12860}$
\end{tcolorbox}

\end{small}
% \vspace{-1mm}
\end{tcolorbox}
}
\caption{Error Case 3 of GPT-o1.}
\label{app:o1_error_math_1}
\end{figure*}

\begin{figure*}[t]
\begin{tcolorbox}[colback=blue!12!white,colframe=blue!75!black,title=Error Case of GPT-o1]
\vspace{-1mm}
\begin{small}
\textbf{Question:}

We write $\lfloor X \rfloor$ to mean the greatest integer less than or equal to $X$; for example $\lfloor 3\frac{1}{2} \rfloor = 3$. If $N = \frac{3}{11}$, what is the value of $\lfloor 10N \rfloor + \lfloor 100N \rfloor + \lfloor 1000N \rfloor + \lfloor 10000N \rfloor + \lfloor 100000N \rfloor + \lfloor 1000000N \rfloor + \lfloor 10000000N \rfloor$?

\textbf{Answer:}

3030299
\begin{tcolorbox}
[colback=red!5!white,colframe=red!75!black,title=O1's wrong answer]
We are given that \( N = \frac{3}{11} \), and we need to calculate the following sum:

\[
\lfloor 10N \rfloor + \lfloor 100N \rfloor + \lfloor 1000N \rfloor + \lfloor 10000N \rfloor + \lfloor 100000N \rfloor + \lfloor 1000000N \rfloor + \lfloor 10000000N \rfloor
\]

Let's calculate each term step by step.

Step 1: Calculate \( N = \frac{3}{11} \)
\[
N = \frac{3}{11} \approx 0.272727\ldots
\]
This is a repeating decimal.

Step 2: Calculate \( \lfloor 10N \rfloor \)
\[
10N = 10 \times 0.272727\ldots = 2.72727\ldots
\]
The greatest integer less than or equal to \( 2.72727\ldots \) is \( \lfloor 10N \rfloor = 2 \).

Step 3: Calculate \( \lfloor 100N \rfloor \)
\[
100N = 100 \times 0.272727\ldots = 27.272727\ldots
\]
The greatest integer less than or equal to \( 27.272727\ldots \) is \( \lfloor 100N \rfloor = 27 \).

Step 4: Calculate \( \lfloor 1000N \rfloor \)
\[
1000N = 1000 \times 0.272727\ldots = 272.727272\ldots
\]
The greatest integer less than or equal to \( 272.727272\ldots \) is \( \lfloor 1000N \rfloor = 272 \).

Step 5: Calculate \( \lfloor 10000N \rfloor \)
\[
10000N = 10000 \times 0.272727\ldots = 2727.272727\ldots
\]
The greatest integer less than or equal to \( 2727.272727\ldots \) is \( \lfloor 10000N \rfloor = 2727 \).

Step 6: Calculate \( \lfloor 100000N \rfloor \)
\[
100000N = 100000 \times 0.272727\ldots = 27272.727272\ldots
\]
The greatest integer less than or equal to \( 27272.727272\ldots \) is \( \lfloor 100000N \rfloor = 27272 \).

Step 7: Calculate \( \lfloor 1000000N \rfloor \)
\[
1000000N = 1000000 \times 0.272727\ldots = 272727.272727\ldots
\]
The greatest integer less than or equal to \( 272727.272727\ldots \) is \( \lfloor 1000000N \rfloor = 272727 \).

Step 8: Calculate \( \lfloor 10000000N \rfloor \)
\[
10000000N = 10000000 \times 0.272727\ldots = 2727272.727272\ldots
\]
The greatest integer less than or equal to \( 2727272.727272\ldots \) is \( \lfloor 10000000N \rfloor = 2727272 \).

Step 9: Add all the terms
Now, we add the results:{\LARGE \textcolor{red}{\XSolidBrush} }\textcolor{red}{(Miscalculation)}
\[
\lfloor 10N \rfloor + \lfloor 100N \rfloor + \lfloor 1000N \rfloor + \lfloor 10000N \rfloor + \lfloor 100000N \rfloor + \lfloor 1000000N \rfloor + \lfloor 10000000N \rfloor
\]
\[
= 2 + 27 + 272 + 2727 + 27272 + 272727 + 2727272
\]
\[
= 3028301
\]

Thus, the value of the expression is:

\[
\boxed{3028301}
\]
\end{tcolorbox}

\end{small}
% \vspace{-1mm}
\end{tcolorbox}
\caption{Error Case 4 of GPT-o1.}
\label{app:o1_error_math_2}
\end{figure*}

\begin{figure*}[!htbp]
\begin{tcolorbox}[colback=blue!12!white,colframe=blue!75!black,title=Prompt]
\vspace{-1mm}
\begin{small}
You are a mathematician with expertise in Algebra, Calculus, Statistics, Geometry, Linear Algebra, and Number Theory, and you are also good at Python coding. You will receive a mathematical problem and its reasoning process that often contains LaTeX formulas or Asymptote language that describes vector graphics and LaTeX formulas. The answer to the problem is wrapped in \verb|\boxed{}|. Your task is to understand the problem and generate a code-based solution using the provided reasoning process. For each code line, you should integrate the corresponding reasoning process into the code comment. Ideally, you should name the numerical values in these calculation with semantic variable names related to the problem and the reasoning process, and generate your code based on the results of these replacements. Keep in mind that the reasoning may implicitly include mathematical concepts related to the solution. Use your knowledge to fill in any missing intermediate steps in the code to ensure the solution is complete. Wrap the generation in <answer></answer> tags.

\bigskip
You should keep in mind that: 

\bigskip
1. Strictly follow the required XML format.

2. For complex outputs (e.g., intervals, complex numbers, coordinates), represent them as strings. Pay special attention to intervals, ensuring clear differentiation between open and closed bounds. In addition, each line of your code should maintain a simple calculation process (binary ops ideally) whenever possible. 

3. When necessary, you can generate your code based on python packages such as `math', `sympy', `scipy', and `numpy'.

4. Ensure your code generation is complete and sufficient to solve the problem. Make sure the running result of the code is exactly same as the answer wrapped in \verb|\boxed{}|.

5. Ensure that the integrated output is coherent. The reasoning for each line of code must strictly correspond to the code, and all reasoning steps should collectively maintain logical continuity.

\bigskip
Here are some examples to show you how to do it:

\bigskip
\verb|{Few Shot Examples}|

\bigskip
How about this question?

\bigskip
Question: \verb|{question}|

\bigskip
Reasoning: \verb|{reasoning}|

\end{small}
% \vspace{-1mm}
\end{tcolorbox}
\caption{Prompt for code generation.}
\label{fig:prompt_code_gen}
\end{figure*}

\begin{figure*}[!htbp]
\begin{tcolorbox}[colback=blue!12!white,colframe=blue!75!black,title=Prompt for GSM8K]
\vspace{-1mm}
\begin{small}

You are given a problem, and its solution process including both reasoning text and corresponding python code. Then I will add more steps in solution code, your tasks are as follows: 

1. Create a new question and the corresponding solution process by filling in the \verb|[REASON_PAD]| placeholder before each line of code with reasoning that directly relates to it.

The added intervention content is indicated in "\verb|extra_var|" and ``\verb|op_var|", and there maybe some intermediate variables started with ``\verb|intermediate_x|" to help you understand how this intervention is resulted. Your question should reflect both the information of "\verb|extra_var|" and the information of "\verb|intermediate_x|" (if any), and they should be causally linked. You can utilize the running results commented after the code line to generate the question and reasoning process, but you should calculate the final result by yourself, which is commented as "?". Return (1) the modified question wrapped in (\verb|<question></question>| tags, (2) the correspond solution process wrapped in <solution></solution> tags, and (3) the final answer in \verb|<answer></answer>| tags. 

2. Rename "\verb|extra_var|" and "\verb|op_var|" based on the modified question and modified reasoning process, wrapped "\verb|extra_var|" in \verb|<exvar></exvar>| tags and "\verb|op_var|" in \verb|opvar></opvar>| tags.

3. Evaluate whether the new question and reasoning process are logically sound and free from factual, computational, or logical errors. Return the evaluation as either \verb|<eval>CORRECT</eval>| or \verb|<eval>INCORRECT</eval>|. 

You should keep in mind that: 

1. Strictly follow the required XML format, and do not alter the code.

2. Ensure the question is clear and precise, eliminating any ambiguity by clearly defining all terms and scenarios.

3. Ensure the integrated output is complete and sufficient to solve the problem.

4. Ensure that the integrated output is coherent. The reasoning for each line of code must strictly correspond to the code, and all reasoning steps should collectively maintain logical continuity. 

5. When evaluating, keep in mind that for a math problem, in the interest of computational simplicity, some exaggerated statements (such as excessive work or a huge amount of food intake, etc.) and professional errors in other fields are permissible. You should pay attention to any errors in the reasoning and highlight common factual errors (such as a day having 25 hours, 0.5 people, eating -2 apples, etc.) if they are not pre-assumed. Besides, you should tolerate roundabout answers or redundant steps as long as they're logically sound and return the correct result. 

\bigskip
Here is one example to show you how to do this task,

\bigskip
\verb|{Few Shot Example}|

\bigskip
How about this question?

\bigskip
\verb|####|

\bigskip
QUESTION: \verb|{question}|

\bigskip
SOLUTION:

\bigskip
```

\verb|{program}|

```

\bigskip
If we add more steps into the code, we will have the new solution program as:

\bigskip
\verb|```|

\verb|{program_intervened}|

\verb|```|

\bigskip
OUTPUT: 

\bigskip
MODIFIED SOLUTION: 

\bigskip
MODIFIED QUESTION: 

\bigskip
ANSWER:

\bigskip
RENAME:

\bigskip
EVALUATION:

\end{small}
% \vspace{-1mm}
\end{tcolorbox}
\caption{Prompt for code translation for GSM8K.}
\label{fig:prompt_code_trans_GSM8K}
\end{figure*}

\begin{figure*}[!htbp]
\begin{tcolorbox}[colback=blue!12!white,colframe=blue!75!black,title=Prompt for MATH]
\vspace{-1mm}
\begin{small}
You are a mathematician with expertise in Algebra, Calculus, Statistics, Geometry, Linear Algebra, and Number Theory. You are given a mathematical problem that may contain LaTeX or Asymptote code to describe tabular data and vector graphics. You are also given its solution code, of which each line represents one step in solving the problem, with comments explaining the reasoning process. Based on the provided problem and code, your task is to 

1. Generate a more challenging problem that requires one additional reasoning step compared to the original problem. Ensure the problem remains logically consistent and relevant. You need to provide (1) the new problem enclosed in \verb|<question></question>| and (2) the complete solution code enclosed in \verb|<code></code>| following the format of the provided solution with comments explaining the reasoning process. Moreover, to verify that an additional step has been incorporated, (3) specify and explain the extra step added to the solution process and enclose it in \verb|<added_code></added_code>|.

2. Evaluate whether the new question and reasoning process are logically sound and free from factual, computational, or logical errors. Return the evaluation as either \verb|<eval>CORRECT</eval>| or \verb|<eval>INCORRECT</eval>|. Note that for the interest of computational simplicity, some exaggerated statements (such as excessive work or a huge amount of food intake, etc.) and professional errors in other fields are permissible. You should pay attention to any reasoning errors and highlight common factual errors (such as a day having 25 hours, 0.5 people, eating -2 apples, etc.) if they are not pre-assumed.
\bigskip
You should keep in mind that: 

1. Strictly follow the required xml format. 

2. Ensure the question is clear and precise, eliminating any ambiguity by clearly defining all terms and scenarios. 

3. Read and understand the latex and asymptote language in the reasoning process, you should keep them in the new question and modify them if necessary.

4. When necessary, you can generate your code based on python packages such as `math', `sympy', `scipy', and `numpy'.

5. Remember to update the names of corresponding variables in the subsequent code to reflect the modification.

6. When evaluating, pay special attention to the occurrence of decimals and negative results, as they often lead to unreasonable results in real-world or mathmatical contexts (such as 0.5 people, eat -3 apples, etc.)
\bigskip
Here are some examples to show you how to do this task:

\bigskip

\verb|{Few Shot Examples}|
\bigskip

How about this question?
\bigskip

\verb|####|

\bigskip
QUESTION: \verb|{question}|

\bigskip
CODE:

\bigskip
\verb|```|

\verb|{program}|

\verb|```|

\bigskip
OUTPUT:

\bigskip
NEW QUESTION:

\bigskip
NEW CODE:

\bigskip
ADDED CODE: 

\bigskip
EVALUATION:

\end{small}
% \vspace{-1mm}
\end{tcolorbox}
\caption{Prompt for code translation for MATH.}
\label{fig:prompt_code_trans_MATH}
\end{figure*}

\begin{figure*}[!htbp]
\begin{tcolorbox}[colback=blue!12!white,colframe=blue!75!black,title=Prompt]
\vspace{-1mm}
\begin{small}
Evaluate if the given question and the corresponding reasoning process are logically sound. Return the evaluation as either \verb|<eval>CORRECT</eval>| or \verb|<eval>INCORRECT</eval>|. Keep in mind that for a math problem, in the interest of computational simplicity, some exaggerated statements (such as excessive work or a huge amount of food intake, etc.) and professional errors in other fields are permissible. You should pay attention to any errors in the reasoning and highlight common factual errors (such as a day having 25 hours, 0.5 people, eating -2 apples, etc.) if they are not pre-assumed. Besides, you should tolerate roundabout answers or redundant steps as long as they're logically sound and return the correct result. Return your answer only. Additionally, if you assess it as incorrect, offer brief explanations enclosed in \verb|<explain></explain>|.

    \bigskip
    QUESTION:  \verb|{question}|
    
    \bigskip
    REASONING:\verb|{reasoning}|

\end{small}
% \vspace{-1mm}
\end{tcolorbox}
\caption{Prompt for evaluating generated problems.}
\label{fig:prompt_gpt_eval}
\end{figure*}

\begin{figure*}[!htbp]
\begin{tcolorbox}[colback=blue!12!white,colframe=blue!75!black,title=Prompt]
\vspace{-1mm}
\begin{small}

Below is a math problem. Please solve it step by step and put your answer in `\verb|\boxed{}|'.

\verb|###|

\bigskip
Problem: \verb|{problem}|

\bigskip
Let's think step by step:

\end{small}
% \vspace{-1mm}
\end{tcolorbox}
\caption{Prompt for training and inferencing.}
\label{fig:prompt_train}
\end{figure*}

\clearpage
\begin{figure*}[t]
\begin{tcolorbox}[colback=blue!12!white,colframe=blue!75!black,title=Error Case of Finetuned Qwen-2-7b]
\vspace{-1mm}
\begin{small}
\textbf{Question:}

The function $f(x)$ satisfies$f(x + y) = f(x) f(y)$for all real numbers $x$ and $y.$  If $f(2) = 9,$ find $f(5).$

\textbf{Answer:}

\begin{tcolorbox}[colback=red!5!white,colframe=red!75!black,title=Correct answer by pretrained Qwen-2-7b]
The given functional equation $f(x + y) = f(x) f(y)$ is a well-known property of exponential functions. In particular, if $f(x)$ is an exponential function of the form $f(x) = a^x$ for some constant $a$, then it satisfies the given functional equation.

To see why, let's substitute $f(x) = a^x$ into the functional equation:

$$f(x + y) = f(x) f(y) \Rightarrow a^{x+y} = a^x a^y \Rightarrow a^{x+y} = a^{x+y}.$$

This equality holds for all real numbers $x$ and $y$, so the functional equation is satisfied by the exponential function $f(x) = a^x$.

Now, we are given that $f(2) = 9$. Since $f(x) = a^x$, we have $a^2 = 9$, which implies $a = 3$ or $a = -3$. However, since $f(x)$ is an exponential function, it must be positive for all real numbers $x$. Therefore, we must have $a = 3$.

Finally, we can find $f(5)$ by substituting $x = 5$ into the exponential function $f(x) = 3^x$:

$$f(5) = 3^5 = 243.$$

Therefore, the value of $f(5)$ is $\boxed{243}$.\LARGE \textcolor{teal}{\Checkmark}
\end{tcolorbox}

\begin{tcolorbox}[colback=red!5!white,colframe=red!75!black,title=Wrong answer by finetuned Qwen-2-7b using our dataset]
Given that f(2) = 9. Let's verify this property with some test values. Let's use x = 2 and y = 2 to demonstrate the functional equation. Calculate f(2) * f(2) = 9 * 9 = 81. Verify that f(4) = f(2 + 2) = f(2) * f(2) = 81. Calculate f(5) = f(2 + 3) = f(2) * f(3). Since f(2) = 9, we need to find f(3). From f(4) = f(2 + 2) = f(2) * f(2) = 81, we can find f(3) = f(4)/f(2) = 81/9 = 9. Therefore, f(5) = f(2) * f(3) = 9 * 9 = 81. \#\#\#\# The answer is $\boxed{81}$.{\LARGE \textcolor{red}{\XSolidBrush} }
\end{tcolorbox}

\end{small}
% \vspace{-1mm}
\end{tcolorbox}
\caption{Illustration of erroneous answer after finetuning.}
\label{fig:appen_cata_forget}
\end{figure*}
\end{document}